\documentclass[journal]{IEEEtran}

\usepackage{epsfig}
\usepackage{graphicx}

\usepackage{algorithmic}
\usepackage{amsmath}
\usepackage{indentfirst}
\usepackage{cite}
\usepackage{amsmath,bm}
\usepackage[linesnumbered,boxed,ruled,commentsnumbered,vlined]{algorithm2e}
\usepackage{algorithmic}
\usepackage{dashrule}

\usepackage{amssymb}
\usepackage{times}
\usepackage{graphicx}
\usepackage{subfigure}
\usepackage{multirow}
\usepackage{multicol}
\usepackage{amsmath}
\newcommand{\tabincell}[2]{\begin{tabular}{@{}#1@{}}#2\end{tabular}}

\usepackage[ruled,linesnumbered]{algorithm2e}

\usepackage{threeparttable}
\usepackage{dcolumn}
\usepackage{multirow}
\usepackage{multicol}
\usepackage{booktabs}
\usepackage{subfigure}

\usepackage{amsthm}

\usepackage{latexsym}

\usepackage{rotating}
\usepackage{textcomp}

\usepackage{colortbl}
\definecolor{light-gray}{gray}{0.75} 

\usepackage{xcolor}
\usepackage[normalem]{ulem} 
\newcommand\hl{\bgroup\markoverwith
  {\textcolor{yellow}{\rule[-.5ex]{2pt}{2.5ex}}}\ULon}

\begin{document}
%
\title{ATM-R: An Adaptive Tradeoff Model with Reference Points for Constrained Multiobjective Evolutionary Optimization
}

\author{\IEEEauthorblockN{Bing-Chuan Wang, Yunchuan Qin, Xian-Bing Meng, Zhi-Zhong Liu}

\thanks{B.-C. Wang is with the School of Automation, Central South University, Changsha 410083, China (email: bingcwang@csu.edu.cn).}
	
\thanks{Y. Qin and Z.-Z. Liu are with the College of Information Science and Electronic Engineering, Hunan University, Changsha 410082, China (e-mail: liuzz@hnu.edu.cn; qinyunchuan@hnu.edu.cn).}

\thanks{X.-B. Meng is with the School of Computer Science and Engineering, South China University of Technology, Guangzhou 510006, China (e-mail: axbmeng@scut.edu.cn).}
}

\maketitle

\begin{abstract}

The goal of constrained multiobjective evolutionary optimization is to obtain a  set of well-converged and well-distributed feasible solutions.
To complete this goal, there should be a tradeoff among feasibility, diversity, and convergence.
However, it is nontrivial to balance these three elements simultaneously by using a single tradeoff model since the importance of each element varies in different evolutionary phases. As an alternative, we adapt different tradeoff models in different phases and propose a novel algorithm called ATM-R.
In the infeasible phase, ATM-R takes the tradeoff between diversity and feasibility into account, aiming to move the population toward feasible regions from diverse search directions.
In the semi-feasible phase, ATM-R promotes the transition from ``the tradeoff between feasibility and diversity'' to ``the tradeoff between diversity and convergence'', which can facilitate the discovering of enough feasible regions and speed up the search for the feasible Pareto optima in succession.
In the feasible phase, the tradeoff between diversity and convergence is considered to attain a set of well-converged and well-distributed feasible solutions.
It is worth noting that the merits of reference points are leveraged in ATM-R to accomplish these tradeoff models.
Also, in ATM-R, a multiphase mating selection strategy is developed to generate promising solutions beneficial to different evolutionary phases.
Systemic experiments on a wide range of benchmark test functions demonstrate that ATM-R is effective and competitive, compared against five state-of-the-art constrained multiobjective optimization evolutionary algorithms.

\end{abstract}


%
\IEEEpeerreviewmaketitle

\section{Introduction}\label{sec:intro}

\IEEEPARstart{M}{any} scientific or engineering problems involve the optimization of conflicting objectives subject to constraints, which can be formulated as constrained multiobjective optimization problems (CMOPs)~\cite{miettinen2012nonlinear}:

\begin{equation}\label{equ:CMOP}
\begin{aligned}
\min \quad &\mathbf{F(x)}=(f_1(\mathbf{x}),f_2(\mathbf{x}),\cdots,f_m(\mathbf{x}))^\text{T}\in\mathbb{R}^m\\
\text{s.t.}\quad & g_j(\mathbf{x})<0,\;j=1,\cdots,n_g\\
& h_j(\mathbf{x})=0,\;j=n_g+1,\cdots,n_g+n_h\\
& \underline{x}_j\leq x_j\leq \overline{x}_j,\;j=1,\cdots,D
\end{aligned},
\end{equation}

\noindent where $\mathbf{F(x)}$ denotes the objective vector consisting of $m$ conflicting objectives (i.e., $f_i(\mathbf{x}),i=1,\cdots,m$); $\mathbf{x}=(x_1,\cdots,x_D)^\text{T}$ is a $D$-dimensional decision vector/solution; $\underline{x}_j$ and $\overline{x}_j$ are the lower and upper bounds of $x_j$, respectively; $\mathbb{S}=\prod_{j=1}^{D}[\underline{x}_j,\overline{x}_j]$ refers to the decision space; $g_j(\mathbf{x})$ and $h_j(\mathbf{x})$ represent the $j$th inequality and $(j-n_g)$th equality constraints, respectively; $n_g$ and $n_h$ are the numbers of the inequality and equality constraints, respectively.
When solving a CMOP, we always quantify constraint violation by the degree of constraint violation:
\begin{equation}\label{equ:CV}
G(\mathbf{x})=\sum_{j=1}^{n_g+n_h}G_j(\mathbf{x}).
\end{equation}
\noindent $G_j(\mathbf{x})$ denotes the degree of constraint violation of the $j$th constraint~\cite{wang2018composite}:
\begin{equation}\label{equ:consvio}
G_{j}(\mathbf{x})=\left\{ \begin{array}{ll} {\rm max}(0,g_{j}(\mathbf{x})), \quad 1 \leq j \leq n_g \\ {\rm max}(0,\left| h_{j}(\mathbf{x})\right|-\delta ), \quad n_g+1 \leq j \leq n_g+n_h \end{array} \right.
\end{equation}
\noindent where $\delta$ is a small positive value used to relax an equality constraint to some degree. A solution $\mathbf{x}$ is called a feasible solution, if and only if $G(\mathbf{x})=0$. All feasible solutions constitute the feasible region: $\Omega=\{\mathbf{x}\in\mathbb{R}^D|G(\mathbf{x})=0\}$.
For two solutions $\mathbf{x}_u, \mathbf{x}_v \in \Omega$, $\mathbf{x}_u$ is said to Pareto dominate $\mathbf{x}_v$, denoted as $\mathbf{x}_u\prec \mathbf{x}_v$, if and only if $\forall j\in\{1,\cdots,m\}, f_j(\mathbf{x}_u)\leq f_j(\mathbf{x}_v)\bigwedge \exists j\in\{1,\cdots,m\}, f_j(\mathbf{x}_u)< f_j(\mathbf{x}_v)$.
A solution $\mathbf{x}_p\in\Omega$ is considered as a Pareto optimum if and only if $\neg\exists \mathbf{x}_v\in\Omega, \mathbf{x}_v\prec\mathbf{x}_p$.
The set of all Pareto optima is called the constrained Pareto set, and its image in the objective space is called the constrained Pareto front (CPF). 
The goal of constrained multiobjective evolutionary optimization is to pursue a  set of well-converged and well-distributed feasible solutions to approximate the CPF.


To complete this goal, a consensus has been reached in the community of constrained multiobjective optimization that a good tradeoff among feasibility, diversity, and convergence should be achieved~\cite{ma2021shift}. It is worth noting that the importance of these three elements varies in different evolutionary phases. Let us take the element of feasibility for example. In the infeasible phase, this element is very important because feasibility information plays an indispensable role in locating feasible regions, which is crucial for constrained multiobjective optimization. However, in the feasible phase, this element is negligible as all solutions become feasible. We only need to consider the tradeoff between diversity and convergence. Due to their varied importance, it is nontrivial to balance these three elements simultaneously by using a single tradeoff model.

\begin{figure}[t]\setlength{\abovecaptionskip}{0.cm}
	\setlength{\belowcaptionskip}{-0pt}
	\centering
	\includegraphics[width=3.5in]{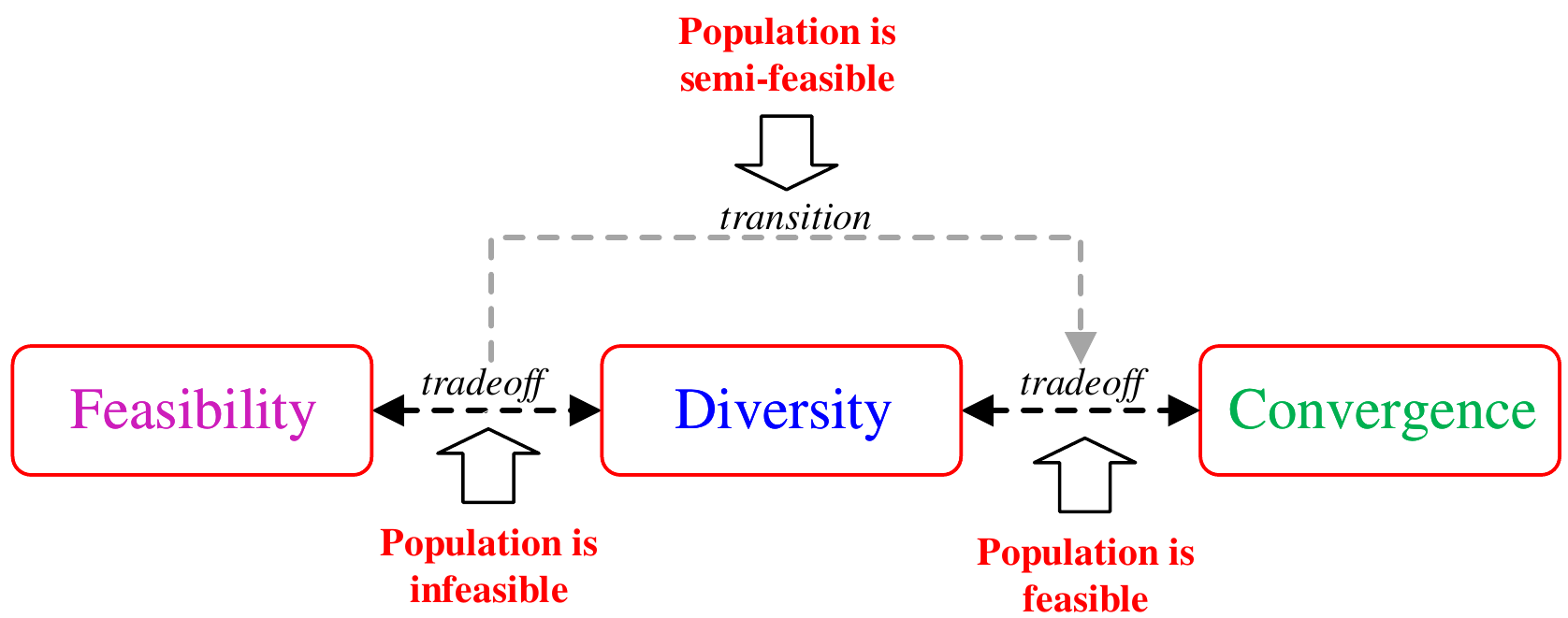}\\
	\caption{Task decomposition of achieving a tradeoff among feasibility, diversity, and convergence.}\label{fig:methodology}
\end{figure}

As an alternative, we adapt different tradeoff models in different evolutionary phases, proposing an \emph{a}daptive \emph{t}radeoff \emph{m}odel with \emph{r}eference points (ATM-R) to handle CMOPs.
Fig.~\ref{fig:methodology} depicts the tradeoffs considered in ATM-R:
\begin{itemize}	
	\item
\emph{achieving a tradeoff between feasibility and diversity in the infeasible phase}:
when the population is entirely infeasible, the primary goal is to find as many feasible regions as possible since the Pareto optima may be scattered in different feasible regions.
To this end, a tradeoff between feasibility and diversity should be achieved to move the population toward the feasible regions from diverse search directions.	
	
	\item
\emph{promoting the transition from ``the tradeoff between feasibility and diversity" to ``the tradeoff between diversity and convergence" in the semi-feasible phase}:
when the population is semi-feasible (i.e., the population contains both infeasible and feasible solutions), two situations should be considered.
In the early stage, only a few feasible regions are discovered.
In this case, the tradeoff between feasibility and diversity should still be prioritized to find more promising feasible regions.
Once enough feasible regions are located, in the later stage, attention should be paid to drive the population toward the CPF quickly and make them uniformly spread over the CPF simultaneously.
Thus, the tradeoff between convergence and diversity should be concentrated on.
In summary, in this phase, we should shift from ``the tradeoff between feasibility and diversity'' to ``the tradeoff between diversity and convergence''~\cite{ma2021shift}.
	
	\item
\emph{achieving a tradeoff between diversity and convergence in the feasible phase}: when the population is completely feasible, the final task is to move the feasible solutions toward the CPF quickly while maintaining good diversity.
Apparently, a tradeoff between diversity and convergence should be realized~\cite{liu2021handling}. 	
\end{itemize}

In summary, the core of a CMOEA is how to accomplish the above tradeoffs.
The tradeoff in the feasible phase has been well studied in the community of evolutionary multiobjective optimization.
For convenience, in ATM-R, an off-the-shelf unconstrained multiobjective optimization evolutionary algorithm (MOEA) is utilized to achieve this tradeoff directly.
As for the tradeoffs in the other two phases, the related studies remain relatively scarce.
Especially for the tradeoff in the semi-feasible phase, little research focuses on this topic.
Indeed, to achieve the tradeoffs in these two phases, an important concern is how to deal with the infeasible solutions.
Past experience in the community of evolutionary constrained multiobjective optimization has shown that the infeasible solutions can not only facilitate maintaining diversity but also contribute to speeding up the convergence.
In ATM-R, the merits of reference points are leveraged to select different kinds of infeasible solutions suitable for different evolutionary phases.
In summary, the main contributions of this paper are as follows:
\begin{itemize}
	\item

   Instead of using a single tradeoff model, we adapt different tradeoff models in different evolutionary phases, proposing a novel constrained multiobjective optimization algorithm (CMOEA) called ATM-R.
   Although it is inevitable for an algorithm to experience three phases during the evolution, few attempts have been made to develop alternate tradeoff models for different phases to facilitate a more explicit adaptation.


	\item

    By leveraging the merits of reference points, we provide a new perspective that selects promising infeasible solutions suitable for different evolutionary phases.
    To the best of our knowledge, relevant work along this direction remains scarce.




	\item

    A multiphase mating selection strategy is developed in this paper that adaptively selects suitable mating parents for different evolutionary phases.

	\item

   Systemic experiments have been implemented on three sets of test suites including 36 benchmark CMOPs to validate the effectiveness of ATM-R.
   Comparison against five state-of-the-art CMOEAs suggests that ATM-R is significantly superior or comparable to the contender algorithms on most of the test problems.
   Additionally, the advantages of some important algorithmic components in ATM-R have been verified.

\end{itemize}

The rest of this paper is organized as follows. Section II conducts a brief review of related CMOEAs. The details of ATM-R are described in Section III. The performance of ATM-R is compared with five representative CMOEAs in Section IV. Section V presents some further analyses of ATM-R in depth. The concluding remarks and future work are given in Section VI.

\section{Related Work}

Constrained multiobjective optimization has become a hot topic in the community of evolutionary computation and numerous CMOEAs have been proposed. Based on whether infeasible solutions are utilized, these CMOEAs can be classified into two categories: feasibility-driven CMOEAs and infeasibility-assisted CMOEAs. 

\subsection{Feasibility-Driven CMOEAs}
A feasibility-driven CMOEA is driven by feasibility information, in which feasible solutions are considered to be better than infeasible ones. Some feasibility-driven CMOEAs use the constrained dominance principle (CDP) to compare two solutions~\cite{deb2002a}. In the CDP, a solution $\mathbf{x}_u$ is said to be better than another solution $\mathbf{x}_v$, if one of the following conditions is met:
\begin{itemize}
	\item both $\mathbf{x}_u$ and $\mathbf{x}_v$ are infeasible, and $G(\mathbf{x}_u)<G(\mathbf{x}_v)$;
	\item $\mathbf{x}_u$ is feasible and $\mathbf{x}_v$ is infeasible;
	\item both $\mathbf{x}_u$ and $\mathbf{x}_v$ are feasible, and $\mathbf{x}_u\prec \mathbf{x}_v$.
\end{itemize}
Due to its preference for feasible solutions, the CDP can motivate the population toward feasible regions quickly. It has been widely integrated with different kinds of MOEAs~\cite{liu2020and,jain2013evolutionary} and used in a spectrum of engineering optimization prob-lems~\cite{ning2017constrained,hobbie2021comparison}. Liu {\it et al.}~\cite{liu2020and} combined an angle-based selection strategy, the shift-based density estimation strategy, and the CDP for constrained many-objective optimization. Jain and Deb~\cite{jain2013evolutionary} proposed a reference-point-based nondominated sorting approach, which is integrated with the CDP for constrained many-objective optimization. Jan and Khanum~\cite{jan2013study} embedded the CDP into the framework of MOEA/D and compared its performance with that of the stochastic ranking~\cite{runarsson2000stochastic}. CDP-based CMOEAs are often used as the baseline algorithms when evaluating the performance of a CMOEA~\cite{fan2019push,ma2019new,peng2017evolutionary}.

The feasibility rule, which is widely used for constrained single-objective optimization, has been extended to solve CMOPs. Liu {\it et al.}~\cite{liu2019indicator-based} combined the feasibility rule with an indicator-based MOEA and compared its performance with that of some other kinds of CMOEAs. Fan {\it et al.}~\cite{fan2017comparative} carried out a comparison study on MOEA/D for constrained multiobjective optimization. Different constraint-handling techniques including the feasibility rule are embedded into the framework of MOEA/D.

Some CMOEAs put emphasis on constraints when the population contains no feasible solutions. Woldesenbet and Yen~\cite{woldesenbet2009constraint} presented a self-adaptive penalty method to solve CMOPs, in which an adaptive penalty function and a distance measure are combined for constraint-handling. In fact, when the population is entirely infeasible, the self-adaptive penalty method compares two solutions based on constraints regardless of objectives. Liu and Wang~\cite{liu2019handling} presented a two-phase CMOEA to solve CMOPs. When the population is entirely infeasible, all objectives are combined together and the feasibility rule is used to tackle constraints.  Due to the superior capability of its search algorithm, the two-phase CMOEA can handle complex constraints in the decision space.  Jimenez {\it et al.}~\cite{jimenez2002evolutionary} designed a CMOEA for constrained multiobjective optimization, in which the min-max formulation is used to tackle constraints. In addition, the feasibility rule is used to compare two solutions when an offspring is inserted into the new population. Miyakawa {\it et al.} ~\cite{miyakawa2013two} developed a two-stage nondominated sorting method to solve CMOPs. The population is divided into several fronts by the nondominated sorting according to constraints. The obtained fronts are further partitioned by the nondominated sorting based on objectives. In this manner, constraints are prior to objectives in environmental selection.

\subsection{Infeasibility-assisted CMOEAs}
An infeasibility-assisted CMOEA takes advantage of infeasible solutions for constrained multiobjective optimization. Most state-of-the-art CMOEAs fall into this category.

Some CMOEAs take advantage of infeasible solutions implicitly by using a comparison criterion that takes both constraints and objectives into account. Ma and Wang~\cite{ma2021shift} proposed a shifted-based penalty function, in which an infeasible solution is penalized based on the information provided by the feasible solutions nearby. Jiao {\it et al.}~\cite{jiao2014modified} proposed a modified objective function method. When the population is entirely infeasible, the modified objective function is equivalent to a distance measure in which constraints and objectives are considered equally important. Fan {\it et al.}~\cite{fan2016angle} presented an angle-based CDP for constrained multiobjective optimization. Given a feasible solution and an infeasible solution, if the angle between these two solutions is smaller than a predefined threshold, they would be nondominated each other. Thus, some infeasible solutions could enter into the new population instead of some feasible ones. Young~\cite{young2005blended} proposed a blended ranking measure to select solutions. By blending an individual's rank in the objective space with its rank in the constraint space, an infeasible solution may be better than a feasible one. Similarly, Ma {\it et al.}~\cite{ma2019new} designed a new fitness function with two rankings, in which one ranking value is obtained based on the CDP and the other is calculated based on the Pareto dominance without considering constraints. The $\varepsilon$ constrained method can use infeasibility information by tuning a threshold value $\varepsilon$~\cite{wang2018composite}; thus, it has been widely used to solve CMOPs~\cite{yang2014epsilon}. Zapotecas-Mart{\'\i}nez and Ponsich~\cite{zapotecas2020constraint} combined MOEA/D with the $\varepsilon$ constrained method to solve CMOPs, in which the $\varepsilon$ value is set according to the degree of constraint violation. Fan {\it et al.}~\cite{fan2016improved} improved the $\varepsilon$ constrained method by setting the $\varepsilon$ value dynamically. Zhou {\it et al.}~\cite{zhou2021infeasible} extended the $\varepsilon$ constrained method to solve CMOPs. When the degree of constraint violation of an infeasible solution is larger than the $\varepsilon$ value, its diversity will be carefully maintained. The stochastic ranking that is popular for constrained single-objective optimization has also been extended to solve CMOPs~\cite{geng2006infeasible,liu2019indicator-based}.

Some CMOEAs leverage the advantages of infeasible solutions explicitly by archiving or coevolution. Ray {\it et al.}~\cite{ray2009infeasibility} proposed an infeasibility-driven EA, in which a small percentage of infeasible solutions close to the constraint boundaries are maintained. Li {\it et al.}~\cite{li2018two} designed a two-archive EA for constrained multiobjective optimization. An archive is used to promote convergence, while the other is used to maintain diversity. The diversity archive evolves without considering constraints; thus, infeasible solutions with good objective function values can be fully used. Liu {\it et al.}~\cite{liu2021handling} tried to solve CMOPs through bidirectional coevolution. The CDP is used to drive the main population toward the CPF from the feasible side of the search space. In addition, a nondominated sorting procedure and an angle-based selection scheme are conducted in sequence to motivate the population toward the CPF within the infeasible region. Tian {\it et al.}~\cite{tian2020coevolutionary} developed a coevolutionary framework for constrained multiobjective optimization. Similarly, one population is updated by the CDP, while the other is updated by an unconstrained MOEA. Additionally, the elites of these two populations are selected to generate offspring. Ishibuchi {\it et al.}~\cite{ishibuchi2018dual} designed a dual-grid model of MOEA/D for constrained multiobjective optimization. Two populations are maintained and infeasible solutions with good objective function values are preferred in the secondary population. Zhu {\it et al.}~\cite{zhu2019moea} employed two types of weight vectors in MOEA/D to solve CMOPs. The solutions associated with the convergence weight vectors are updated based on the aggregation function, while the solutions associated with the diversity weight vectors are renewed according to both the aggregation function and the degree of constraint violation. Peng {\it et al.}~\cite{peng2017evolutionary} used two kinds of weight vectors for constrained multiobjective optimization. Specifically, the degree of constraint violation is considered as another objective. Subsequently, a set of feasible weight vectors and a set of infeasible weight vectors are used to update the population. Additionally, the set of infeasible weight vectors is dynamically adjusted to maintain a number of infeasible solutions with good objective function values and small degrees of constraint violation.

Some CMOEAs divide the evolutionary process into several phases and put emphasis on objectives in one of the phases. Yang {\it et al.}~\cite{yang2020constrained} divided the evolutionary process into a constrained search mode and an unconstrained search mode. These two search modes are executed by a dynamic constraint-handling mechanism. Fan {\it et al.}~\cite{fan2019push} proposed a push and pull search (PPS) framework to solve CMOPs, in which the evolutionary process is divided into two stages: push and pull. In the push stage, the population is updated by an unconstrained MOEA. In the pull stage, an improved $\varepsilon$ constrained method is designed to tackle complex constraints. Since its proposition, the PPS framework has been used in various fields~\cite{fan2020push,peng2021constrained}. Yu {\it et al.}~\cite{yu2021dynamic} proposed a dynamic selection preference-assisted constrained multiobjective differential evolutionary (DE) algorithm. The selection preference for a solution shifts from infeasibility to feasibility as the optimization progresses. Tian {\it et al.}~\cite{tian2020balancing} proposed a two-stage CMOEA to balance objective optimization and constraint sanctification. These two stages are executed dynamically according to the percentage of feasible solutions in the population. Recently, Ming {\it et al.}~\cite{MING2021107263} proposed a simple two-stage EA for constrained multiobjective optimization. The two-stage EA focuses on approaching the unconstrained Pareto front in the first stage and the feasible solutions are archived. In the second stage, the method seeks to approximate the CPF, where the archived feasible solutions are adopted as the initial population. Peng {\it et al.}~\cite{peng2020cooperative} proposed a two-phase EA for constrained multiobjective optimization with deceptive constraints. In the first phase, two subpopulations are employed to explore the feasible regions and the entire space, respectively. The second phase aims to approach the CPF. Additionally, an infeasibility utilization strategy is designed to leverage the promising information provided by infeasible solutions.

\section{Proposed Method}

\begin{figure}[t]\setlength{\abovecaptionskip}{0.cm}
	\setlength{\belowcaptionskip}{-0pt}
	\centering
	\includegraphics[width=4in]{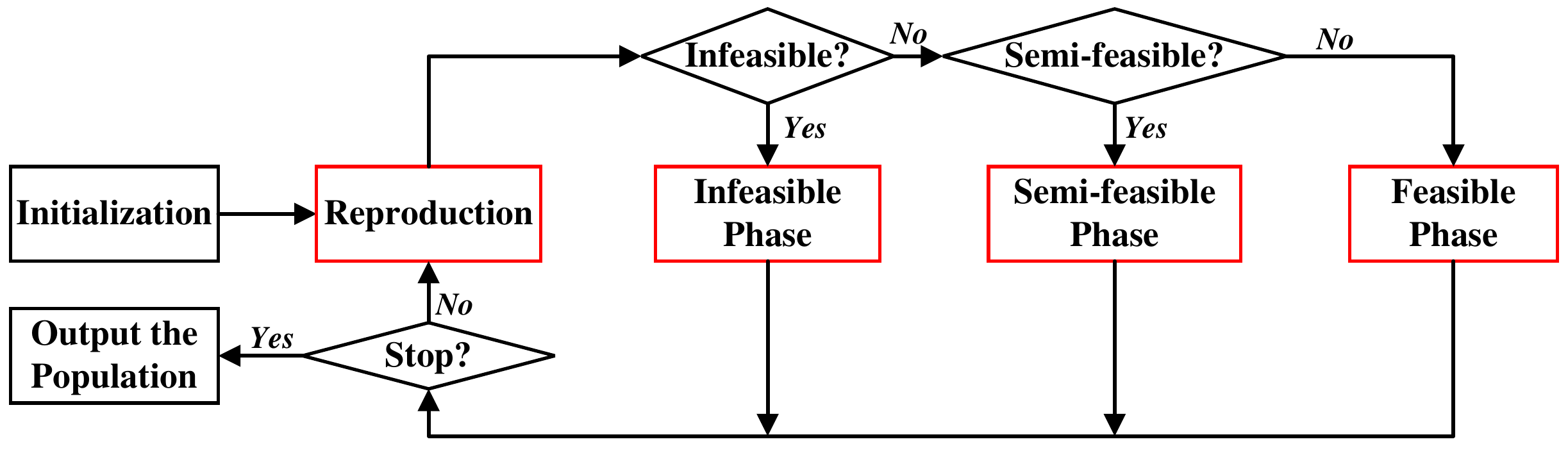}\\
	\caption{Flow chart of ATM-R.}\label{fig:Framework}
\end{figure}

The general flow chart of ATM-R is shown in Fig.~\ref{fig:Framework}.
As its name implies, ATM-R makes use of reference points to adaptively accomplish different tradeoffs in different evolutionary phases, those are, the infeasible phase, the semi-feasible phase, and the feasible phase.
The details of the update mechanisms in these three different phases are described in Section~\ref{sec:infeasible}, Section~\ref{sec:semi-feasible}, and Section~\ref{sec:feasible}, respectively.
Aside from the environmental selection procedure, another critical element of a CMOEA is the mating selection procedure.
In ATM-R, a multiphase mating selection strategy is developed to generate promising solutions beneficial to different tradeoffs.
The details of this strategy are illustrated in Section~\ref{sec:reproduction}. Finally, the framework of ATM-R and some discussions are shown in Section~\ref{sec:method} and Section~\ref{sec:discussion}, respectively.

\begin{algorithm}[tp]
\caption{Update Mechanism in the Infeasible Phase}\label{alg:infeasible}
\KwIn{Population $\mathbb{P}$, offspring population $\mathbb{O}$}
\KwOut{New population $\mathbb{P}$}
$\mathbb{Q}\leftarrow \mathbb{P}\cup\mathbb{O}$\;
Divide $\mathbb{Q}$ into $k$ fronts based on $\mathbf{\hat{F}(x)}$: $\mathbb{F}_1,\cdots,\mathbb{F}_k$\;
$\mathbb{P}\leftarrow\varnothing$\;
\For{$l=1:k$}{
   \If{$|\mathbb{P}|+|\mathbb{F}_l|\geq N$}{
      Break\;
    }
   $\mathbb{P}\leftarrow \mathbb{P}\cup\mathbb{F}_l$\;
}
\If{$|\mathbb{P}|+|\mathbb{F}_l| > N$}
  {
  	Sample $n$ uniformly distributed reference points and generate corresponding weight vectors: $\mathbf{w}_1,\cdots,\mathbf{w}_n$\;
  	Assign each solution in $\mathbb{F}_l$ to a weight vector according to \eqref{equ:assign}-\eqref{equ:normalize}\;
  	\While{$|\mathbb{P}|+|\mathbb{F}_l| > N$}
  	      {
  	      	Select the weight vector associated with the largest number of solutions: $\mathbf{w}_c$\;
  	      	Among the solutions assigned to $\mathbf{w}_c$, select the one with the largest value of $G(\mathbf{x})$: $\mathbf{x}_w$\;
  	      	$\mathbb{F}_l\leftarrow \mathbb{F}_l \backslash \mathbf{x}_w$\;
  	      }
  }

  $\mathbb{P}\leftarrow \mathbb{P}\cup\mathbb{F}_l$\;

\end{algorithm}

\subsection{Update Mechanism in the Infeasible Phase}\label{sec:infeasible}

In this phase, ATM-R aims to strike a balance between feasibility and diversity.
In other words, it motivates the population toward feasibility from diverse search directions, thus locating as many feasible regions as possible.
\textbf{Algorithm~\ref{alg:infeasible}} shows how ATM-R accomplishes this tradeoff.
In general, it involves two essential elements.

\subsubsection{Nondominated Sorting in the Transformed Objective Space}

Following the ideas in~\cite{wang2008adaptive}, we consider $G(\mathbf{x})$ as an additional objective function, and transform~\eqref{equ:CMOP} into an unconstrained MOP:
\begin{equation}\label{eqn:tCMOP}
\text{min}~\mathbf{\hat{F}(x)}=(f_1(\mathbf{x}),\cdots,f_m(\mathbf{x}),G(\mathbf{x}))^\text{T}\in\mathbb{R}^{m+1}.
\end{equation}
Clearly, this transformation does not introduce any extra parameters.
In addition, both objective functions and constraints are considered in~\eqref{eqn:tCMOP}, which can facilitate maintaining population diversity and enhance driving forces toward the feasible regions.
Based on $\mathbf{\hat{F}(x)}$, the population will be divided into several fronts, denoted as $\mathbb{F}_1,\cdots,\mathbb{F}_k$, by implementing a nondominated sorting procedure in the transformed objective space.
Afterward, the solutions in each front will be selected in turn until $\sum_{i=1}^{l-1}|\mathbb{F}_i|<N\leq \sum_{i=1}^{l}|\mathbb{F}_i|$ where $N$ denotes the size of the final solution set.

\subsubsection{Regular Reference Point-based Selection}

If $\sum_{i=1}^{l}|\mathbb{F}_i|$ is larger than $N$, we should further select $(n=N-\sum_{i=1}^{l-1}|\mathbb{F}_i|)$ solutions from the last desired front $\mathbb{F}_l$.
To complete this task, in this study, a regular reference point-based selection scheme is developed by taking advantage of uniformly distributed reference points.
Its implementation is quite simple.
\begin{itemize}
\item
First, a set of regular (i.e., uniformly distributed) reference points is sampled in the objective space to generate weight vectors denoted as $\{\mathbf{w}_1,\cdots,\mathbf{w}_n\}$ following the ideas in~\cite{das1998normal}.
\item
Subsequently, a solution (denoted as $\mathbf{x}$) in $\mathbb{F}_l$ is assigned to the weight vector with the smallest angle to its normalized objective vector:
\begin{equation}\label{equ:assign}
I=\mathop{\arg\min}_{j\in\{1,\cdots,n\}}\theta_{j},
\end{equation}
\begin{equation}\label{equ:theta}
\theta_{j}=\text{arccos}\left|\frac{\mathbf{F'(x)}^\text{T}\mathbf{w}_j}{\|\mathbf{F'(x)}\|\cdot\|\mathbf{w}_j\|}\right|,\;j=1,\cdots,n,
\end{equation}
\begin{equation}\label{equ:normalize}
f_j'(\mathbf{x})=\frac{f_j(\mathbf{x})-z_j^{\text{min}}}{z_j^{\text{max}}-z_j^{\text{min}}},\; j=1,\cdots,m,
\end{equation}
where $I$ indicates which weight vector the solution $\mathbf{x}$ is assigned to; $\theta_j$ denotes the angle between $\mathbf{w}_j$ and the normalized objective vector $\mathbf{F'(x)}=(f_1'(\mathbf{x}),\cdots,f_m'(\mathbf{x}))^\text{T}$;
$\|\cdot\|$ represents the function to calculate the 2-norm of a vector; $\mathbf{z}^{\text{max}}=(z_1^{\text{max}},\cdots,z_m^{\text{max}})^\text{T}$ and $\mathbf{z}^{\text{min}}=(z_1^{\text{min}},\cdots,z_m^{\text{min}})^\text{T}$ refer to the estimated nadir point and ideal point, respectively.
\item
Afterward, $(|\mathbb{F}_l|-n)$ inferior solutions are deleted one by one by employing a ``diversity first, feasibility second'' strategy.
To be specific, it first identifies the weight vector associated with the largest number of solutions\footnote{Note that the tie is broken at random}.
Intuitively, since these solutions are associated with the same weight vector, they will share highly similar search directions.
To maintain diverse search directions, it is necessary to delete one of them.
The feasibility information of these solutions is considered for the deletion. The one with the largest value of $G(\mathbf{x})$  is discarded.
These two steps will continue until $(|\mathbb{F}_l|-n)$ solutions are deleted.
\end{itemize}

\begin{figure}[t]\setlength{\abovecaptionskip}{0.cm}
	\setlength{\belowcaptionskip}{-0pt}
	\centering
	\includegraphics[width= 5.5cm]{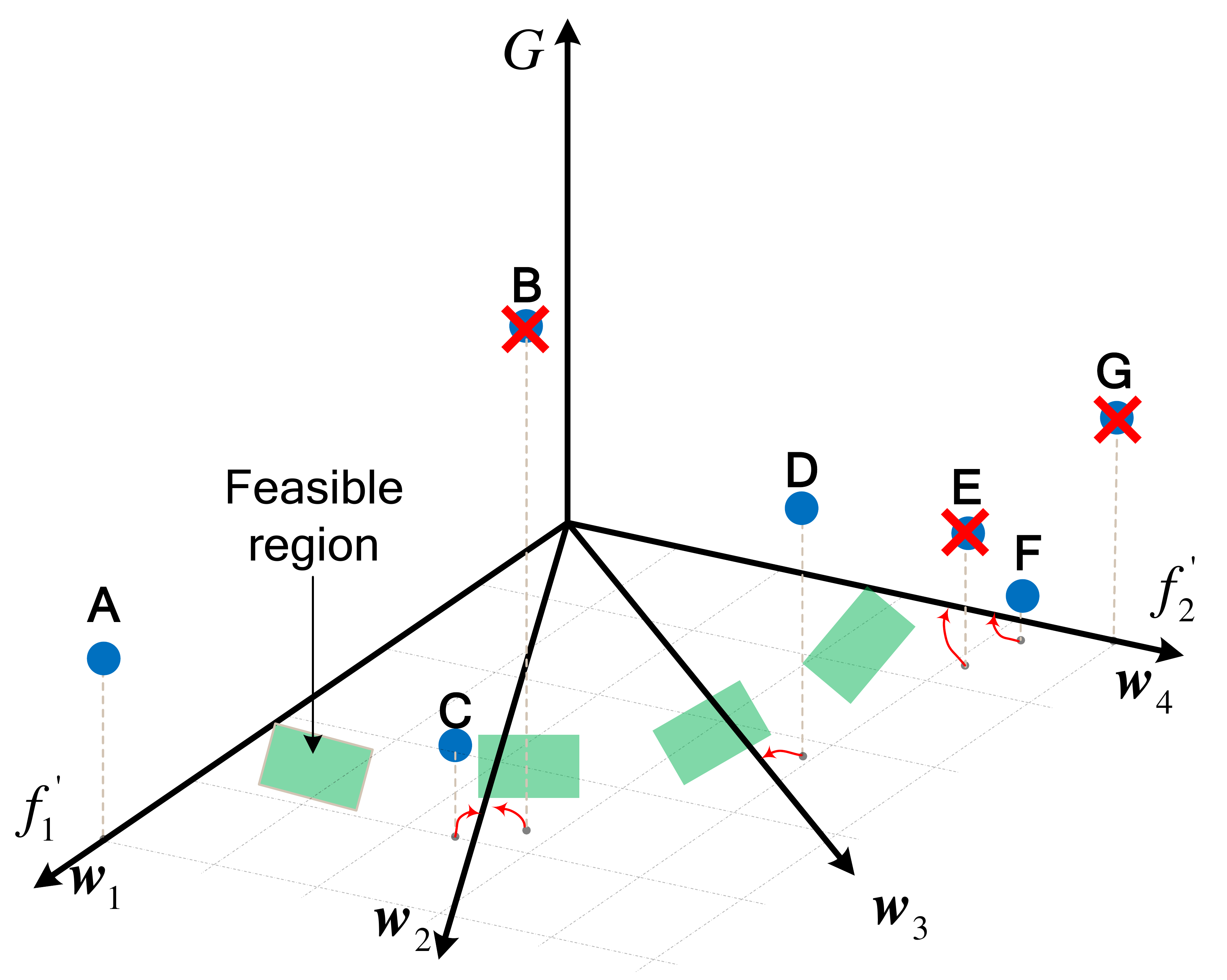}\\
	\caption{Update mechanism in the infeasible phase.}\label{fig:infeasible}
\end{figure}

A simple example is given in Fig.~\ref{fig:infeasible} for better understanding the regular reference point-based selection scheme.
We consider a CMOP with two objectives.
Suppose there are seven solutions in the population, and they lie in the same front in the transformed objective space.
According to the values of $G(\mathbf{x})$, these individuals were ranked as $\textbf{F}$, $\textbf{C}$, $\textbf{E}$, $\textbf{A}$, $\textbf{G}$, $\textbf{D}$, and $\textbf{B}$ in ascending order.
The task is to select four solutions for the next generation.
\begin{enumerate}
	\item
   First, four reference points are sampled uniformly to generate four weight vectors denoted as $\{\mathbf{w}_1,\cdots,\mathbf{w}_4\}$.
	\item
    Next, each solution in the population is assigned to a weight vector: $\mathbf{w}_1\leftrightarrow\{\textbf{A}\}$, $\mathbf{w}_2\leftrightarrow\{\textbf{B},\textbf{C}\}$, $\mathbf{w}_3\leftrightarrow\{\textbf{D}\}$, and $\mathbf{w}_4\leftrightarrow\{\textbf{E},\textbf{F},\textbf{G}\}$.
	\item
    Subsequently, three solutions are deleted one by one.
    $\textbf{G}$ is first deleted since $\mathbf{w}_4$ is matched with the largest number of solutions and $\textbf{G}$ is the one with the largest value of $G(\mathbf{x})$ compared with $\textbf{E}$ and $\textbf{F}$.
    According to this principle, $\textbf{B}$ and $\textbf{E}$ will be also removed.
	\item
    Finally, the solutions (i.e.,~\textbf{A},~\textbf{C},~\textbf{D}, and~\textbf{F} will enter into the next generation.
\end{enumerate}

\emph{Remark 1}: Both ATMES\footnote{Although ATMES is originally designed for constrained single-objective optimization, it can be directly applied to solve CMOPs.}~\cite{wang2008adaptive} and IDEA~\cite{ray2009infeasibility} employ nondominated sorting in the transformed objective space as ATM-R does.
The main difference lies in how to distinguish the solutions in the same front. Specifically, in ATMES, solutions are selected based on $G(\mathbf{x})$ only. A solution with a smaller value of $G(\mathbf{x})$ will be preferred. In this manner, ATMES will put too much emphasis on constraints. It will cause performance deterioration in terms of the search diversity, which is essential for finding as many promising feasible regions as possible. On the contrary, in IDEA, only the diversity in the transformed objective space is considered to update the last desired front $\mathbb{F}_l$.
Unfortunately, this manner will result in a limited driving force toward the feasible regions, which in turn leads to a relatively low convergence speed. Unlike these two methods, ATM-R takes both diversity and feasibility into account to update $\mathbb{F}_l$, and a ``diversity first, feasibility second'' strategy is thus developed. As illustrated in Fig.~\ref{fig:infeasible}, ATM-R can strike a good balance between diversity and feasibility, thereby motivating the population toward feasible regions from diverse search directions.

\subsection{Update Mechanism in the Semi-feasible Phase} \label{sec:semi-feasible}

ATM-R intends to promote the transition from ``the tradeoff between feasibility and diversity'' to ``the tradeoff between diversity and convergence'' in the semi-feasible phase (i.e., the population contains both infeasible and feasible solutions). The reasons for this transition are two-fold. In the early stage of the semi-feasible phase, ATM-R must locate as many feasible regions as possible. To this end, it must focus on the tradeoff between feasibility and diversity. After finding a sufficient number of feasible regions, in the later stage, ATM-R should steer the population rapidly toward the CPF and distribute it uniformly along with the CPF simultaneously. Thus, the tradeoff between convergence and diversity should be prioritized. \textbf{Algorithm~\ref{alg:semi-feasible}} shows how ATM-R updates the solutions in the semi-feasible phase.

\begin{algorithm}[tp]
	\caption{Update Mechanism in the Semi-feasible Phase}\label{alg:semi-feasible}
	\KwIn{Population $\mathbb{P}$, offspring population $\mathbb{O}$, $FEs$, $MaxFEs$}
	\KwOut{New population $\mathbb{P}$}
	$\mathbb{Q}\leftarrow \mathbb{P}\cup\mathbb{O}$, $\mathbb{P}\leftarrow\varnothing$\;
	$\mathbb{Q}_f\leftarrow\{\mathbf{x}\in\mathbb{Q}|G(\mathbf{x})=0\}$,
	$\mathbb{Q}_{if}\leftarrow\{\mathbf{x}\in\mathbb{Q}|G(\mathbf{x})>0\}$\;
	
	\If{$|\mathbb{Q}_f|>N$}{
		$\mathbb{Q}_f\leftarrow$ $N$ feasible solutions seleted from $\mathbb{Q}_f$ by an unconstrained MOEA\;	
	}
	$\mathbb{P}\leftarrow \mathbb{P}\cup\mathbb{Q}_f$\;

	\If{$|\mathbb{Q}_{if}|>N$}
	{
		\uIf{$\frac{FEs}{MaxFEs}<0.5\;\text{or}\;|\mathbb{Q}_{f}|<N$}
		{
			$\mathbb{Q}_{if}\leftarrow$ $N$ infeasible solutions selected from $\mathbb{Q}_{if}$ by using \textbf{Algorithm~\ref{alg:infeasible}}\;	
		}
		\Else
		{
			Generate $|\mathbb{Q}_f|$ weight vectors by using the solutions in $\mathbb{Q}_f$ according to \eqref{equ:weight}-\eqref{equ:normalize1}\;
			Assign each solution in $|\mathbb{Q}_{if}|$ to a weight vector according to \eqref{equ:assign}-\eqref{equ:normalize}\;
			\While{$|\mathbb{Q}_{if}|>N$}
			{
				Select the weight vector associated with the largest number of solutions: $\mathbf{w}_c$\;
				Among the solutions assigned to $\mathbf{w}_c$, select the one furthest from the feasible solution used to generate $\mathbf{w}_c$: $\mathbf{x}_w$\;
				$\mathbb{Q}_{if}\leftarrow \mathbb{Q}_{if} \backslash \mathbf{x}_w$\;
			}
			
		}
	}
	$\mathbb{P}\leftarrow \mathbb{P}\cup\mathbb{Q}_{if}$\;
	
\end{algorithm}
From~\textbf{Algorithm~\ref{alg:semi-feasible}}, it is observed that ATM-R updates the feasible and infeasible solutions separately. To update the feasible solutions, an unconstrained MOEA is used to truncate the feasible population $\mathbb{Q}_f$ if its size is greater than $N$;
otherwise, all feasible solutions are reserved. To update the infeasible solutions, ATM-R considers two situations. In the early stage, it aims to achieve a tradeoff between feasibility and diversity, which is the same as in the infeasible phase. Thus, the update mechanism used in the infeasible phase (i.e., \textbf{Algorithm 1}) can be directly applied in this stage. While in the later stage, ATM-R shifts the emphasis to the tradeoff between diversity and convergence.
To realize this tradeoff, an important task is how to preserve those infeasible solutions that can contribute to both diversity and convergence.
ATM-R designs the following two steps to accomplish this task.

\subsubsection{Discovery of the Nondominated Infeasible Solutions}

Compared with the feasible solutions in the current population, the nondominated infeasible solutions usually have smaller objective function values. It is natural to leverage their benefits to promote convergence. To distinguish these infeasible solutions, we first employ a nondominated sorting procedure to divide the union population (i.e., $\mathbb{Q}$ in \textbf{Algorithm~\ref{alg:semi-feasible}}) into several fronts based on $\mathbf{\hat{F}(x)}$ in ~\eqref{eqn:tCMOP}. Subsequently, the infeasible solutions in the first front are picked out. If the number of these nondominated infeasible solutions (denoted as $M$) is smaller than $N$, all of them will be kept; otherwise, they will be further distinguished by the following adaptive reference point-based selection.

\subsubsection{Adaptive Reference Point-based Selection}

\begin{figure}[!t]
    \begin{center}
        \subfigure[]{\label{fig:regular}\includegraphics[width=0.6\columnwidth]{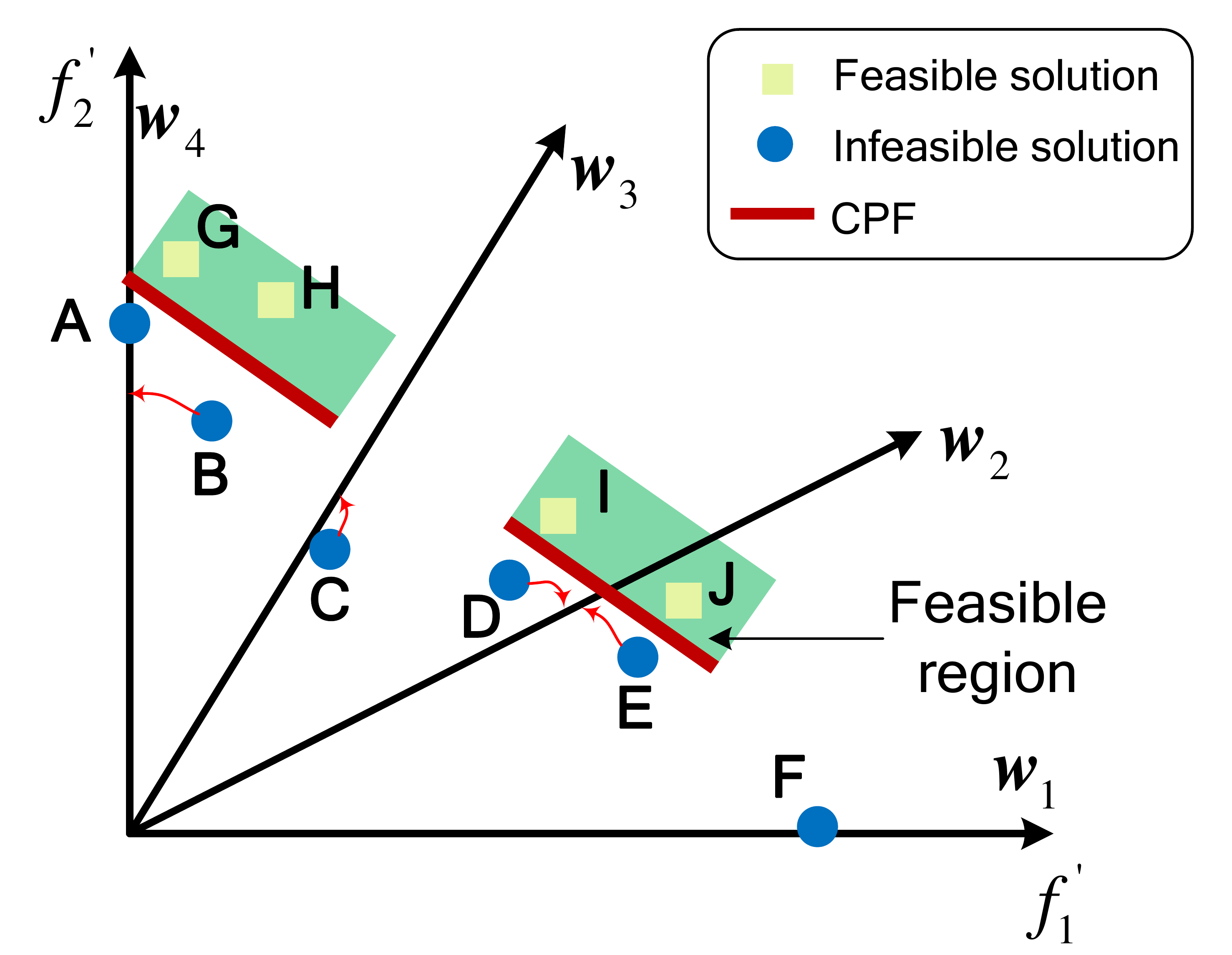}}
        \subfigure[]{\label{fig:adaptive}\includegraphics[width=0.6\columnwidth]{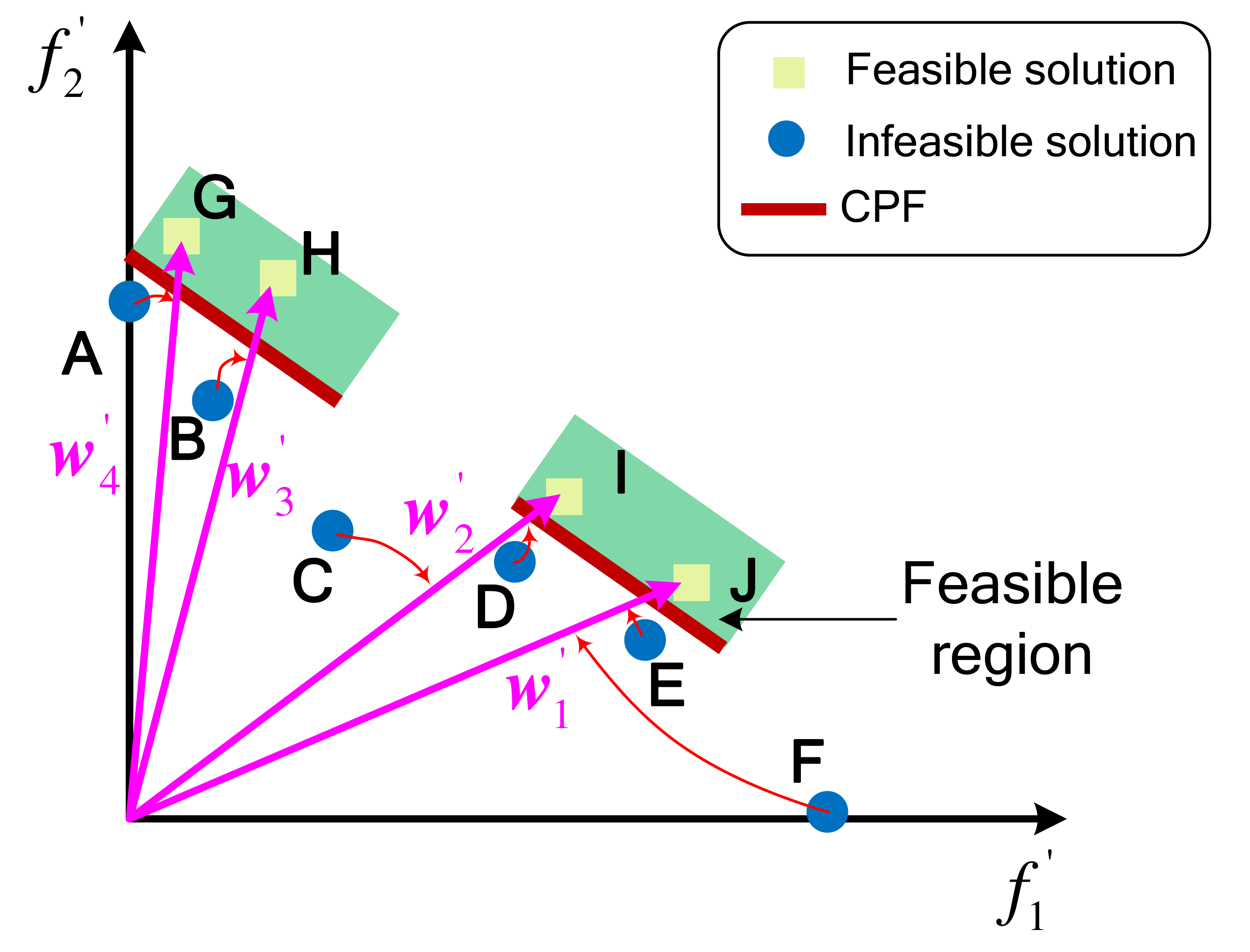}}
        \caption{Illustration of difference between the weight vectors in the regular reference point-based selection and those in the adaptive reference point-based selection: (a) weight vectors in regular reference point-based selection and (b) weight vectors in adaptive reference point-based selection. }\label{fig:difference}
    \end{center}
\end{figure}

Herein, the regular reference points are no longer used to assist the selection.
The reason is that the CPF might be disconnected (see Fig.~\ref{fig:difference}), and some weight vectors (i.e., $\mathbf{w}_1$ and $\mathbf{w}_3$ in Fig.~\ref{fig:regular}) generated using the uniformly distributed reference points cannot  point to any parts of the CPF. As a result, the solutions preserved by making use of such weight vectors (i.e., \textbf{C} and \textbf{F}) are far away from the CPF and hardly contribute to convergence speed, which is not desirable. Instead, we use adaptive reference points for solution selection. For convenience, in our study, the feasible solutions are considered as adaptive reference points since they can deliver important clues for the localization of the CPF (see Fig.~\ref{fig:adaptive}). Based on these reference points, a set of adaptive weight vectors can be obtained conveniently.
To be specific, for the $i$th feasible solution $\mathbf{x}_i$, the corresponding weight vector (denoted as $\mathbf{w}^{'}_i=(w^{'}_{i,1},\cdots,w^{'}_{i,m})^\text{T}$) is generated as follows:
\begin{equation}\label{equ:weight}
w^{'}_{i,j}=\frac{f_j^{'}(\mathbf{x})}{\sum_{j=1}^{m}f_j^{'}(\mathbf{x})},\;j=1,\cdots,m,
\end{equation}
\begin{equation}\label{equ:normalize1}
f_j'(\mathbf{x})=\frac{f_j(\mathbf{x})-z_j^{\text{min}}}{z_j^{\text{max}}-z_j^{\text{min}}},\; j=1,\cdots,m,
\end{equation}
where $(f_1'(\mathbf{x}),\cdots,f_m'(\mathbf{x}))^\text{T}$ is the normalized objective vector,  $(z_1^{\text{max}},\cdots,z_m^{\text{max}})^\text{T}$ and $(z_1^{\text{min}},\cdots,z_m^{\text{min}})^\text{T}$ denote the estimated nadir point and ideal point, respectively.
Fig.~\ref{fig:difference} shows the difference between the weight vectors generated using regular reference points and those using adaptive reference points.
It is evident that the weight vectors obtained using adaptive reference points fit better to the characteristics of the CPF.


Once the adaptive weight vectors are prepared, the next procedures in the adaptive reference point-based selection scheme are quite simple.
First, each nondominated infeasible solution is assigned to a weight vector following the ideas in the regular reference point-based selection scheme (see Eqs.~\eqref{equ:assign}-\eqref{equ:normalize}).
Afterward, ($M$-$N$) infeasible solutions are deleted one by one in a two-step manner.
The first step is to identify the weight vector associated with the largest number of solutions.
In the second step, among the solutions assigned to this weight vector, the one furthest from the feasible solution corresponding to the weight vector in the objective space will be deleted.
In general, the first step is similar to many decomposition-based approaches and can help to maintain population diversity.
As for the second step, it can help to retain those infeasible solutions close to the feasible solutions and thus offer a driving force toward the CPF from the infeasible side of the search space.
Intuitively, this way can speed up the convergence.

\emph{Remark 2:} In the semi-feasible phase, the population size in ATM-R is larger than or equal to $N$.
The reason is that a larger population can enhance population diversity, which is critical to both ``the tradeoff between feasibility and diversity" and ``the tradeoff between diversity and convergence''. As for how to determine whether the algorithm has entered the later stage of the semi-feasible phase, we considered two simple conditions which should be satisfied simultaneously. The first condition is that $\frac{FEs}{MaxFEs}$ should be larger than 0.5. 
Note that $FEs$ and $MaxFEs$ denote the function evaluations and the maximum function evaluations, respectively. The second condition relies on the number of feasible solutions which should be equal to $N$. The first condition implies that enough search efforts have been devoted to finding feasible regions, while the second condition is set to ensure a sufficient number of reference points. In the later stage of the semi-infeasible phase, if no nondominated infeasible solutions are discovered, the algorithm will enter the feasible phase.

\subsection{Update Mechanism in the Feasible Phase} \label{sec:feasible}

In this phase, all solutions are feasible.
Under this condition, only the tradeoff between diversity and convergence should be considered, thus motivating the feasible solutions toward the CPF quickly while maintaining good diversity. Apparently, a current effective unconstrained MOEA can be applied to achieve this balance.
Thus, in ATM-R, an off-the-shelf unconstrained MOEA is employed in this phase straightforwardly.


\begin{algorithm}[t]
	\caption{Multiphase Mating Selection Strategy}\label{alg:MPMS}
	\KwIn{Population $\mathbb{P}$, $N$}
	\KwOut{Mating population $\mathbb{C}$}
	
	$\mathbb{C}\leftarrow \varnothing$\;
	
	\For{$i=1:N$}{
	 Randomly select two different solutions denoted as $\mathbf{x}_a$ and $\mathbf{x}_b$ from $\mathbb{P}$\;
	
     \uIf{$\mathbb{P}$ is entirely infeasible}{
     	\uIf{$rand<0.5$}
     	{
     		$\mathbf{x}_m\leftarrow$ the better one between $\mathbf{x}_a$ and $\mathbf{x}_b$ based on the degree of constraint violation\;    	   	
     	}
        \Else
        {
        	$\mathbf{x}_m\leftarrow$ the better one between $\mathbf{x}_a$ and $\mathbf{x}_b$ based on the diversity\;
        }   		
     }
     \uElseIf{$\mathbb{P}$ is feasible}{
     	
     	\uIf{$\mathbf{x}_a\prec \mathbf{x}_b$}
     	{
     		$\mathbf{x}_m\leftarrow\mathbf{x}_a$\;	
     	}
     	\uElseIf{$\mathbf{x}_b\prec \mathbf{x}_a$}
     	{
     		$\mathbf{x}_m\leftarrow\mathbf{x}_b$\;		
     	}
     	\Else
     	{
     		$\mathbf{x}_m\leftarrow$ the better one between $\mathbf{x}_a$ and $\mathbf{x}_b$ based on the diversity\;
     	}     	
     }	
      \ElseIf{$\mathbb{P}$ is semi-feasible}
      {
        \uIf{$i<N/2$}
          {
          	$\mathbf{x}_m\leftarrow$ the better one between $\mathbf{x}_a$ and $\mathbf{x}_b$ by using the method in the infeasible phase\;
          }
        \Else
          {
          	$\mathbf{x}_m\leftarrow$ the better one between $\mathbf{x}_a$ and $\mathbf{x}_b$ by using the method in the feasible phase\;
          }
      }

     $\mathbb{C}\leftarrow\mathbb{C}\cup \mathbf{x}_m$\;	

    }
	
\end{algorithm}

\subsection{Multiphase Mating Selection Strategy} \label{sec:reproduction}
In addition to the multi-phase strategy in environmental selection, ATM-R uses a multi-phase strategy for mating selection. It selects appropriate mating parents  suitable for different evolutionary phases.
The details of this multiphase mating selection strategy are described in  \textbf{Algorithm \ref{alg:MPMS}}.
Similarly, three different phases are considered in this strategy.
\begin{itemize}
\item 
In the infeasible phase, population diversity and feasibility should be focused on simultaneously.
Thus, in the tournament selection, solutions are compared based on the diversity and the degree of constraint violation with the same probability (i.e., 0.5). 
Note that the diversity is quantified by the same way as in~\cite{tian2020coevolutionary}.

\item
In the feasible phase, population diversity and convergence should be taken into account.
Following the ideas in NSGA-II~\cite{deb2002a}, in the tournament selection, solutions are compared based on the Pareto dominance relationship.
Also, if two solutions do not dominate each other, they are compared based on the diversity.
\item
The semi-feasible phase needs to bridge the gap between the feasible phase and the infeasible phase. 
Thus, in this phase, the first half of the mating population is selected by using the method in the infeasible phase, while the other half is selected by using the method in the feasible phase.
\end{itemize}

\begin{algorithm}[t]
	\caption{ATM-R}\label{alg:framework}
	\KwIn{A CMOP, $N$, $MaxFEs$}
	\KwOut{Final population $\mathbb{P}$}
	$\mathbb{P}\leftarrow$ a population initialized from the decision space\;
	$FEs\leftarrow N$\;
	\While{$FEs<MaxFEs$}
	{
		$\mathbb{C}\leftarrow$ a mating population selected from $\mathbb{P}$ by using \textbf{Algorithm~\ref{alg:MPMS}}\;
		$\mathbb{O}\leftarrow$ an offspring population generated by executing genetic operators on  $\mathbb{C}$\;
		$FEs\leftarrow FEs+N$\;
		$\mathbb{Q}\leftarrow \mathbb{P}\cup\mathbb{O}$\;
		
		\uIf{$\mathbb{Q}$ is entirely infeasible}{
			$\mathbb{P}\leftarrow$ the solutions seleted from $\mathbb{Q}$ by using \textbf{Algorithm~\ref{alg:infeasible}}\;	
		}
		\uElseIf{$\mathbb{Q}$ is semi-feasible}
		{
			$\mathbb{P}\leftarrow$ the solutions selected from $\mathbb{Q}$ by using \textbf{Algorithm~\ref{alg:semi-feasible}}\;		
		}
		\ElseIf{$\mathbb{Q}$ is feasible}{
			$\mathbb{P}\leftarrow$ the solutions selected from $\mathbb{Q}$ by using an unconstrained MOEA\;	
		}		
	}		
\end{algorithm}

\subsection{ATM-R}\label{sec:method}
In summary, the details of ATM-R are given in \textbf{Algorithm~\ref{alg:framework}}. 
At the beginning, a population of $N$ solutions is sampled uniformly in the decision space (Lines 1-2). 
Afterward, the population is employed to search for the CPF until the maximum number of function evaluations is exhausted (Lines 3-15). 
In the search process, first, $N$ mating parents are selected for offspring generation by using the multiphase mating selection strategy in \textbf{Algorithm~\ref{alg:MPMS}} (Line 4).
Next, $N$ offspring are produced by the simulated binary crossover (SBX)~\cite{deb1995simulated} and the polynomial mutation (PM)~\cite{deb1996combined} (Lines 5-6). 
Afterward, promising solutions are selected based on population feasibility (Lines 7-14) where \textbf{Algorithm~\ref{alg:infeasible}} and \textbf{Algorithm~\ref{alg:semi-feasible}} are used in the infeasible phase and the semi-feasible phase, respectively. 
Note that if $FEs\geq MaxFEs$, the final population $\mathbb{P}$ would be output.

\subsection{Discussion}\label{sec:discussion}
In essence, ATM-R is a multiphase CMOEA.
ATM-R intends to achieve a tradeoff between diversity and feasibility in the infeasible phase, promote the transition from ``the tradeoff between feasibility and diversity'' to ``the tradeoff between diversity and convergence'' in the semi-feasible phase, and accomplish the tradeoff between diversity and convergence in the feasible phase.
To the best of our knowledge, ATM-R is the first algorithm considering these tradeoffs simultaneously during different evolution phases.
Also, ATM-R is interesting in that it selects promising infeasible solutions suitable for different evolutionary phases by using two kinds of reference points.
As far as we know, relevant studies in this direction are almost absent.
From our analysis, it is apparent that ATM-R is a brand-new CMOEA for constrained multiobjective optimization.

The computational time complexity of ATM-R is mainly determined by the nondominated sorting and the unconstrained MOEA. 
Suppose the fast nondominated sorting and NSGAII~\cite{deb2002a} are adopted in ATM-R.  In the worst case of the infeasible phase, no solutions nondominated another in the transformed objective space.
The time complexity of this nondominated sorting is $O((m+1)\cdot N^2)$.
The time complexity of assigning each solution to a weight vector is $O(m\cdot N^2)$. 
The time complexity of selecting $N$ solutions is $O(N^2)$. 
Thus, the time complexity of the infeasible phase is $O((m+1)\cdot N^2)+O(m\cdot N^2)+O(N^2)=O((m+1)\cdot N^2)$. 
In the semi-feasible phase, the worst-case time complexity of selecting feasible solutions is $O(m\cdot N^2)$. In the early stage of the semi-feasible phase, the worst-case time complexity is the same as that of the infeasible phase: $O((m+1)\cdot N^2)$. 
In the worst case of the later stage, no infeasible solutions nondominated another. It is the same as that of the infeasible phase. Thus, its time complexity is $O((m+1)\cdot N^2)$. 
The time complexity of the semi-feasible phase is $O(m\cdot N^2)+O((m+1)\cdot N^2)+O((m+1)\cdot N^2)=O((m+1)\cdot N^2)$. 
In the feasible phase, the time complexity is the same as that of NSGAII: $O(m\cdot N^2)$. In summary, the computational time complexity of ATM-R is $O((m+1)\cdot N^2)+O((m+1)\cdot N^2)+O(m\cdot N^2)=O(m\cdot N^2)$, which is indeed acceptable.

\section{Performance Comparison}
In this section, we assess the performance of ATM-R based on a wide range of benchmark test functions. Specifically, ATM-R was used to solve three test suites and its performance was compared with that of five representative CMOEAs. Note that all experiments were implemented by the PlatEMO toolbox~\cite{tian2017platemo}.

\subsection{Experimental Settings}
\subsubsection{Test Functions}
Three test suites  consisting of 36 benchmark test functions (e.g., MW~\cite{ma2019evolutionary}, CTP~\cite{deb2001constrained}, and LIRCMOP~\cite{fan2016improved}) were adopted in our study. These test functions own various challenging characteristics; thus, they can assess the performance of a CMOEA adequately. Most state-of-the-art CMOEAs adopt these test functions for empirical study. Note that the number of decision variables in MW and LIRCMOP was set to 15 and 10, respectively. Please see \cite{fan2016improved,ma2019evolutionary,deb2001constrained} for the details of these test functions.

\begin{table*}[ht]
	\centering\scriptsize
	\caption{The IGD values of NSGAII-CDP, PPS, CTAEA, CCMO, ToP, and ATM-R on three sets of benchmark test functions.}
	\begin{tabular}{c c c c c c c}
		\toprule
		Test Functions & \tabincell{c}{NSGAII-CDP\\ mean IGD (std)} & \tabincell{c}{PPS\\ mean IGD (std)}   & \tabincell{c}{CTAEA\\ mean IGD (std)} & \tabincell{c}{CCMO\\ mean IGD (std)}  & \tabincell{c}{ToP\\ mean IGD (std)}   & \tabincell{c}{ATM-R\\ mean IGD (std)} \\
		\midrule
		MW1   & 4.0545e-2 (1.02e-1) - & 2.3190e-2 (4.01e-2) - & 2.1884e-3 (9.96e-4) - & \cellcolor[rgb]{ .647,  .647,  .647}{1.8990e-3 (1.42e-3) +} & NaN (NaN) - & 2.1748e-3 (1.70e-3) \\
		
		MW2   & 2.3926e-2 (7.65e-3) - & 4.2401e-2 (3.25e-2) - & \cellcolor[rgb]{ .647,  .647,  .647}{1.7953e-2 (6.74e-3) $\approx$} & 2.1515e-2 (8.20e-3) - & 2.3108e-1 (1.89e-1) - & 1.9130e-2 (9.76e-3) \\
		
		MW3   & 7.4318e-2 (2.31e-1) - & 7.5935e-3 (9.94e-4) - & 5.4804e-3 (4.86e-4) $\approx$ & \cellcolor[rgb]{ .647,  .647,  .647}{5.2178e-3 (4.41e-4) $\approx$} & 5.9698e-1 (2.78e-1) - & 5.3646e-3 (4.08e-4) \\
		
		MW4   & 5.5780e-2 (2.97e-3) - & 5.3955e-2 (1.73e-3) - & 4.6413e-2 (4.99e-4) - & 4.1285e-2 (3.48e-4) $\approx$ & NaN (NaN) - & \cellcolor[rgb]{ .647,  .647,  .647}{4.1255e-2 (3.45e-4)} \\
		
		MW5   & 4.2761e-1 (3.35e-1) - & 1.4507e-1 (1.97e-1) - & 1.5758e-2 (3.38e-3) - & 4.6474e-3 (7.30e-3) - & NaN (NaN) - & \cellcolor[rgb]{ .647,  .647,  .647}{4.0638e-3 (1.06e-2)} \\
		
		MW6   & 8.0099e-2 (1.51e-1) - & 1.0037e-1 (1.61e-1) - & \cellcolor[rgb]{ .647,  .647,  .647}{1.1188e-2 (6.68e-3) $\approx$} & 5.2473e-2 (1.26e-1) - & 1.0872e+0 (1.81e-1) - & 1.5369e-2 (8.69e-3) \\
		
		MW7   & 1.0205e-1 (1.93e-1) - & 2.5520e-2 (1.88e-2) - & 7.2156e-3 (5.22e-4) - & \cellcolor[rgb]{ .647,  .647,  .647}{4.8994e-3 (4.80e-4) +} & 4.7226e-1 (2.39e-1) - & 5.2004e-3 (4.73e-4) \\
		
		MW8   & 6.1793e-2 (8.78e-3) - & 7.4112e-2 (2.69e-2) - & 5.5531e-2 (2.47e-3) - & 4.9189e-2 (1.58e-2) $\approx$ & 9.5949e-1 (2.01e-1) - & \cellcolor[rgb]{ .647,  .647,  .647}{4.6368e-2 (5.74e-3)} \\
		
		MW9   & 2.1737e-1 (3.10e-1) - & 7.4032e-2 (1.81e-1) - & \cellcolor[rgb]{ .647,  .647,  .647}{8.8691e-3 (9.23e-4) $\approx$} & 5.1927e-2 (1.79e-1) - & NaN (NaN) - & 9.9563e-3 (2.88e-3) \\
		
		MW10  & 2.3341e-1 (2.36e-1) - & 1.3321e-1 (1.48e-1) - & \cellcolor[rgb]{ .647,  .647,  .647}{1.7599e-2 (1.22e-2) $\approx$} & 4.2867e-2 (2.54e-2) - & NaN (NaN) - & 2.7242e-2 (2.33e-2) \\
		
		MW11  & 4.7335e-1 (3.24e-1) - & 1.3565e-2 (2.15e-2) - & 1.6564e-2 (2.80e-3) - & 6.3416e-3 (5.30e-4) $\approx$ & 9.2141e-1 (1.37e-1) - & \cellcolor[rgb]{ .647,  .647,  .647}{6.1791e-3 (2.34e-4)} \\
		
		MW12  & 8.2766e-2 (2.23e-1) - & 2.9552e-2 (1.20e-1) + & \cellcolor[rgb]{ .647,  .647,  .647}{8.0645e-3 (6.84e-4) +} & 3.0553e-2 (1.40e-1) $\approx$ & NaN (NaN) - & 7.8769e-2 (2.24e-1) \\
		
		MW13  & 2.0642e-1 (2.77e-1) - & 1.3735e-1 (6.32e-2) - & \cellcolor[rgb]{ .647,  .647,  .647}{3.8211e-2 (2.66e-2) $\approx$} & 8.2172e-2 (4.41e-2) - & 8.3328e-1 (5.33e-1) - & 5.2527e-2 (3.23e-2) \\
		
		MW14  & 1.2974e-1 (1.27e-2) - & 2.5313e-1 (9.28e-2) - & 1.1279e-1 (6.91e-3) + & \cellcolor[rgb]{ .647,  .647,  .647}{9.8349e-2 (2.41e-3) +} & 4.9059e-1 (5.81e-1) - & 1.1492e-1 (4.72e-2) \\
		\midrule
		
		CTP1  & 8.1699e-2 (6.62e-2) - & 1.9234e-2 (1.80e-2) - & 1.8672e-2 (3.64e-2) - & 4.4317e-3 (1.05e-3) - & 3.9400e-3 (1.48e-4) - & \cellcolor[rgb]{ .647,  .647,  .647}{3.2367e-3 (7.38e-5)} \\
		
		CTP2  & 2.4408e-3 (1.89e-3) - & 3.7003e-3 (6.96e-4) - & 4.6860e-2 (1.25e-2) - & 1.6836e-3 (1.65e-4) - & 4.4453e-3 (1.08e-3) - & \cellcolor[rgb]{ .647,  .647,  .647}{1.4735e-3 (5.94e-5)} \\
		
		CTP3  & 6.2833e-2 (9.95e-2) - & 3.1094e-2 (4.07e-3) - & 5.8093e-2 (5.79e-3) - & 2.2180e-2 (2.24e-3) - & 3.2847e-2 (6.99e-3) - & \cellcolor[rgb]{ .647,  .647,  .647}{1.0066e-2 (1.60e-3)} \\
		
		CTP4  & 2.4494e-1 (1.29e-1) - & 1.4930e-1 (1.84e-2) - & 1.5350e-1 (1.92e-2) - & 1.3538e-1 (2.18e-2) - & 1.8414e-1 (3.29e-2) - & \cellcolor[rgb]{ .647,  .647,  .647}{7.9556e-2 (1.13e-2)} \\
		
		CTP5  & 7.2574e-3 (2.92e-3) - & 1.8168e-2 (6.11e-3) - & 1.8209e-2 (4.61e-3) - & 7.6639e-3 (1.76e-3) - & 1.2167e-2 (3.45e-3) - & \cellcolor[rgb]{ .647,  .647,  .647}{3.3142e-3 (4.13e-4)} \\
		
		CTP6  & 1.1404e-2 (4.04e-4) - & 1.3061e-2 (7.81e-4) - & 3.8535e-2 (5.23e-3) - & 1.0141e-2 (3.56e-4) - & 1.5214e-2 (2.78e-3) - & \cellcolor[rgb]{ .647,  .647,  .647}{9.7103e-3 (3.13e-4)} \\
		
		CTP7  & 1.6882e-3 (1.39e-3) - & 1.6825e-3 (7.14e-5) - & 1.6364e-3 (1.31e-4) - & 1.1669e-3 (4.52e-5) $\approx$ & 1.5176e-3 (5.54e-5) - & \cellcolor[rgb]{ .647,  .647,  .647}{1.1599e-3 (4.62e-5)} \\
		
		CTP8  & 1.2019e-1 (1.45e-1) - & 1.1932e-2 (5.26e-3) - & 3.4505e-2 (4.79e-3) - & 5.5516e-3 (6.49e-4) - & 8.0925e-2 (1.38e-1) - & \cellcolor[rgb]{ .647,  .647,  .647}{4.7357e-3 (2.32e-4)} \\
		\midrule
		
		LIRCMOP1 & 2.6010e-1 (8.10e-2) - & 1.1024e-1 (3.40e-2) - & 3.7900e-1 (1.66e-1) - & 2.0503e-1 (6.82e-2) - & 1.2547e-1 (1.36e-1) - & \cellcolor[rgb]{ .647,  .647,  .647}{3.5295e-2 (1.26e-2)} \\
		
		LIRCMOP2 & 1.9890e-1 (7.22e-2) - & 7.3024e-2 (2.79e-2) - & 1.2324e-1 (6.12e-2) - & 1.1419e-1 (3.19e-2) - & 6.8227e-2 (5.38e-2) - & \cellcolor[rgb]{ .647,  .647,  .647}{3.1146e-2 (9.36e-3)} \\
		
		LIRCMOP3 & 2.4894e-1 (8.38e-2) - & 1.7697e-1 (5.80e-2) - & 3.4751e-1 (1.14e-1) - & 2.0960e-1 (7.98e-2) - & 3.6351e-1 (5.72e-2) - & \cellcolor[rgb]{ .647,  .647,  .647}{2.3380e-2 (1.06e-2)} \\
		
		LIRCMOP4 & 2.3080e-1 (6.23e-2) - & 1.4996e-1 (5.59e-2) - & 2.7661e-1 (1.38e-1) - & 1.9069e-1 (7.18e-2) - & 3.2442e-1 (5.76e-2) - & \cellcolor[rgb]{ .647,  .647,  .647}{2.3928e-2 (1.11e-2)} \\
		
		LIRCMOP5 & 7.3176e-1 (4.81e-1) - & 8.4362e-2 (2.44e-2) - & 1.3918e-1 (4.41e-2) - & 1.4046e-2 (8.21e-3) $\approx$ & 1.2091e-1 (3.48e-1) - & \cellcolor[rgb]{ .647,  .647,  .647}{1.3635e-2 (6.48e-3)} \\
		
		LIRCMOP6 & 5.7447e-1 (4.57e-1) - & 9.5258e-2 (6.31e-2) - & 1.3633e-1 (1.13e-1) - & 1.1357e-2 (7.75e-3) $\approx$ & \cellcolor[rgb]{ .647,  .647,  .647}{6.4593e-3 (3.45e-4) +} & 5.7790e-2 (1.51e-1) \\
		
		LIRCMOP7 & 1.7441e-2 (1.32e-2) $\approx$ & 5.6488e-2 (5.84e-2) - & 2.5246e-2 (9.12e-3) - & 1.1404e-2 (6.38e-3) $\approx$ & \cellcolor[rgb]{ .647,  .647,  .647}{8.6357e-3 (2.52e-4) +} & 1.2400e-2 (5.03e-3) \\
		
		LIRCMOP8 & 3.6946e-2 (4.51e-2) - & 6.6479e-2 (7.05e-2) - & 3.6096e-2 (6.64e-2) - & 9.1531e-3 (5.01e-3) $\approx$ & \cellcolor[rgb]{ .647,  .647,  .647}{8.6820e-3 (4.53e-4) +} & 9.4267e-3 (3.85e-3) \\
		
		LIRCMOP9 & 5.3564e-1 (1.24e-1) - & 1.4063e-1 (8.94e-2) $\approx$ & 1.1622e-1 (5.42e-2) $\approx$ & \cellcolor[rgb]{ .647,  .647,  .647}{3.4398e-2 (3.91e-2) +} & 2.4115e-1 (1.73e-1) - & 1.1216e-1 (7.48e-2) \\
		
		LIRCMOP10 & 3.6496e-1 (9.66e-2) - & 8.2848e-3 (1.47e-2) - & 6.0919e-2 (6.50e-2) - & \cellcolor[rgb]{ .647,  .647,  .647}{5.4399e-3 (3.36e-4) +} & 5.4878e-3 (2.21e-4) + & 6.9018e-3 (6.27e-4) \\
		
		LIRCMOP11 & 2.4114e-1 (1.80e-1) - & 8.1119e-3 (7.48e-3) - & 1.3778e-1 (3.83e-2) - & \cellcolor[rgb]{ .647,  .647,  .647}{2.4538e-3 (8.89e-5) +} & 1.2447e-1 (6.37e-2) - & 5.3691e-3 (1.45e-2) \\
		
		LIRCMOP12 & 1.5180e-1 (8.66e-2) - & 1.5216e-2 (2.43e-2) $\approx$ & 3.1152e-2 (1.68e-2) - & \cellcolor[rgb]{ .647,  .647,  .647}{4.6113e-3 (2.58e-3) $\approx$} & 2.9104e-2 (5.29e-2) $\approx$ & 7.8014e-3 (7.28e-3) \\
		
		LIRCMOP13 & 2.3757e-1 (3.69e-1) - & 1.1968e-1 (3.45e-3) - & 1.0834e-1 (3.97e-4) - & 9.3972e-2 (1.13e-3) - & 1.2450e-1 (3.78e-3) - & \cellcolor[rgb]{ .647,  .647,  .647}{9.3120e-2 (9.31e-4)} \\
		
		LIRCMOP14 & 2.0248e-1 (2.93e-1) - & 1.1859e-1 (3.97e-3) - & 1.1126e-1 (7.98e-4) - & 9.5773e-2 (7.40e-4) - & 1.1883e-1 (4.04e-3) - & \cellcolor[rgb]{ .647,  .647,  .647}{9.4848e-2 (7.79e-4)} \\
		\midrule
		+/-/$\approx$ & 0/35/1 & 1/33/2 & 2/27/7 & 6/19/11 & 4/25/1 &  \\
		\bottomrule
	\end{tabular}%
	\label{tab:performance-IGD}%
\end{table*}%

\subsubsection{Peer Algorithms}
For performance comparison, five representative CMOEAs were taken into consideration: NSGAII-CDP~\cite{deb2002a}, PPS~\cite{fan2019push}, the constrained two-archive EA (CTAEA)~\cite{li2018two}, the coevolutionary constrained  multiobjective optimization (CCMO)~\cite{tian2020coevolutionary}, and the two-phase EA (ToP)~\cite{liu2019handling}. NSGAII-CDP is a classic CMOEA that is usually adopted as a baseline algorithm, while the other four CMOEAs are state-of-the-art algorithms proposed recently. NSGAII-CDP is a feasibility-driven CMOEA and the others are infeasibility-assisted CMOEAs. Among these four infeasibility-assisted CMOEAs, CTAEA and CCMO are multi-population methods which take advantage of infeasible solutions explicitly by an archive or an additional population. PPS and ToP are multiphase methods which divide the evolutionary process into several phases and put emphasis on objectives in some phases. Note that ATM-R is also a multiphase method. 

\subsubsection{Performance Metrics}
Two frequently used performance metrics were adopted to assess the performance of a CMOEA: inverted generational distance (IGD) and hyper-volume (HV). Both IGD and HV can measure the convergence and coverage of a solution set. More details of these two metrics can be found in~\cite{li2019quality}. 

\subsubsection{Parameter Settings}
The parameters involved in the experiments are given as follows:
\begin{itemize}
	\item Size of the final solution set: $N=100$ for all comparison CMOEAs;
	\item $MaxFEs$: $MaxFEs=60,000$ for the MW and CTP test suites, and $MaxFEs=300,000$ for the LIRCMOP test suite;
	\item Number of independent runs: 30.
\end{itemize}

The SBX and PM were used as genetic operators in all CMOEAs except ToP. The parameters of SBX and PM are as follows:
\begin{itemize}
	\item Crossover probability of SBX: 1;
	\item Mutation probability of PM: $1/D$;
	\item Distribution index of SBX and PM: 20.
\end{itemize}

In addition, the algorithm-specific parameters of the five peer CMOEAs were obtained from their original papers.


\begin{table*}[ht]
	\centering\scriptsize
	\caption{The HV values of NSGAII-CDP, PPS, CTAEA, CCMO, ToP, and ATM-R on three sets of benchmark test functions.}
	\begin{tabular}{c c c c c c c}
		\toprule
		Test Functions & \tabincell{c}{NSGAII-CDP\\ mean HV (std)} & \tabincell{c}{PPS\\ mean HV (std)}   & \tabincell{c}{CTAEA\\ mean HV (std)} & \tabincell{c}{CCMO\\ mean HV (std)}  & \tabincell{c}{ToP\\ mean HV (std)}   & \tabincell{c}{ATM-R\\ mean HV (std)} \\
		\midrule
		MW1   & 4.5445e-1 (8.09e-2) - & 4.6529e-1 (3.63e-2) - & 4.8849e-1 (2.03e-3) - & \cellcolor[rgb]{ .647,  .647,  .647}{4.8927e-1 (3.04e-3) +} & NaN (NaN) - & 4.8853e-1 (3.60e-3) \\
	
		MW2   & 5.4798e-1 (1.15e-2) - & 5.2241e-1 (4.44e-2) - & \cellcolor[rgb]{ .647,  .647,  .647}{5.5765e-1 (1.14e-2) $\approx$} & 5.5199e-1 (1.30e-2) - & 3.2482e-1 (1.46e-1) - & 5.5635e-1 (1.54e-2) \\
		
		MW3   & 5.0168e-1 (1.37e-1) - & 5.4398e-1 (4.88e-4) + & \cellcolor[rgb]{ .647,  .647,  .647}{5.4413e-1 (6.14e-4) +} & 5.4368e-1 (7.81e-4) + & 1.2745e-1 (1.27e-1) - & 5.4292e-1 (7.86e-4) \\
		
		MW4   & 8.2309e-1 (5.63e-3) - & 8.2478e-1 (2.49e-3) - & 8.3814e-1 (4.04e-4) - & \cellcolor[rgb]{ .647,  .647,  .647}{8.4116e-1 (4.35e-4) +} & NaN (NaN) - & 8.4001e-1 (7.93e-4) \\
	
		MW5   & 1.7725e-1 (9.80e-2) - & 2.5212e-1 (6.86e-2) - & 3.1449e-1 (2.61e-3) - & 3.2205e-1 (5.38e-3) - & NaN (NaN) - & \cellcolor[rgb]{ .647,  .647,  .647}{3.2214e-1 (6.45e-3)} \\
		
		MW6   & 2.8267e-1 (4.89e-2) - & 2.5928e-1 (6.08e-2) - & \cellcolor[rgb]{ .647,  .647,  .647}{3.1251e-1 (9.93e-3) $\approx$} & 2.9009e-1 (5.16e-2) - & 1.2194e-2 (2.75e-2) - & 3.0911e-1 (1.20e-2) \\
		
		MW7   & 3.7706e-1 (6.78e-2) $\approx$ & 4.0647e-1 (2.09e-3) - & 4.0868e-1 (1.03e-3) - & \cellcolor[rgb]{ .647,  .647,  .647}{4.1205e-1 (5.95e-4) +} & 1.9015e-1 (7.70e-2) - & 4.1019e-1 (9.75e-4) \\
		
		MW8   & 4.9733e-1 (2.20e-2) - & 4.7275e-1 (5.64e-2) - & 5.2198e-1 (1.16e-2) - & 5.2798e-1 (3.48e-2) - & 4.6501e-2 (7.81e-2) - & \cellcolor[rgb]{ .647,  .647,  .647}{5.3338e-1 (1.72e-2)} \\
		
		MW9   & 2.6792e-1 (1.71e-1) $\approx$ & 3.4455e-1 (1.00e-1) - & \cellcolor[rgb]{ .647,  .647,  .647}{3.9100e-1 (2.43e-3) +} & 3.7160e-1 (1.01e-1) - & NaN (NaN) - & 3.8287e-1 (4.60e-3) \\
		
		MW10  & 3.1175e-1 (1.18e-1) - & 3.5982e-1 (7.57e-2) - & \cellcolor[rgb]{ .647,  .647,  .647}{4.3564e-1 (1.30e-2) $\approx$} & 4.1378e-1 (1.88e-2) - & NaN (NaN) - & 4.2764e-1 (1.94e-2) \\
		
		MW11  & 3.2816e-1 (8.07e-2) - & 4.4157e-1 (9.48e-3) - & 4.4127e-1 (1.39e-3) - & 4.4609e-1 (2.03e-3) - & 2.2321e-1 (4.19e-2) - & \cellcolor[rgb]{ .647,  .647,  .647}{4.4746e-1 (2.05e-4)} \\
		
		MW12  & 5.4172e-1 (1.81e-1) - & 5.8181e-1 (1.06e-1) + & \cellcolor[rgb]{ .647,  .647,  .647}{6.0052e-1 (7.80e-4) +} & 5.8415e-1 (1.10e-1) $\approx$ & NaN (NaN) - & 5.4377e-1 (1.82e-1) \\
		
		MW13  & 4.0153e-1 (5.63e-2) - & 4.1137e-1 (4.30e-2) - & \cellcolor[rgb]{ .647,  .647,  .647}{4.6130e-1 (1.23e-2) $\approx$} & 4.3974e-1 (2.53e-2) - & 2.3054e-1 (1.15e-1) - & 4.5371e-1 (1.66e-2) \\
		
		MW14  & 4.5123e-1 (5.66e-3) - & 4.2008e-1 (2.54e-2) - & 4.6575e-1 (3.90e-3) $\approx$ & \cellcolor[rgb]{ .647,  .647,  .647}{4.7246e-1 (1.53e-3) +} & 3.4138e-1 (1.53e-1) - & 4.6217e-1 (1.49e-2) \\
		\midrule
		CTP1  & 3.5920e-1 (1.97e-2) - & 3.7510e-1 (5.31e-3) - & 3.7588e-1 (1.03e-2) - & 3.8065e-1 (3.93e-4) - & 3.8036e-1 (1.15e-4) - & \cellcolor[rgb]{ .647,  .647,  .647}{3.8106e-1 (1.09e-4)} \\
		
		CTP2  & 4.3083e-1 (1.66e-3) - & 4.2928e-1 (7.86e-4) - & 3.9367e-1 (8.22e-3) - & 4.3073e-1 (2.94e-4) - & 4.2689e-1 (1.40e-3) - & \cellcolor[rgb]{ .647,  .647,  .647}{4.3128e-1 (2.45e-4)} \\
		
		CTP3  & 3.7267e-1 (5.45e-2) - & 3.8376e-1 (4.03e-3) - & 3.5588e-1 (7.25e-3) - & 3.9219e-1 (2.18e-3) - & 3.8119e-1 (6.97e-3) - & \cellcolor[rgb]{ .647,  .647,  .647}{4.0507e-1 (1.75e-3)} \\
		
		CTP4  & 2.2838e-1 (6.88e-2) - & 2.5406e-1 (1.86e-2) - & 2.4611e-1 (2.00e-2) - & 2.7265e-1 (2.41e-2) - & 2.2246e-1 (2.88e-2) - & \cellcolor[rgb]{ .647,  .647,  .647}{3.3430e-1 (1.20e-2)} \\		
		
		CTP5  & 3.9329e-1 (2.59e-2) - & 3.8822e-1 (4.53e-3) - & 3.5629e-1 (9.01e-3) - & 3.9643e-1 (2.52e-3) - & 3.8643e-1 (5.29e-3) - & \cellcolor[rgb]{ .647,  .647,  .647}{4.0665e-1 (1.86e-3)} \\
		
		CTP6  & 4.6359e-1 (3.75e-4) - & 4.6198e-1 (6.48e-4) - & 4.4896e-1 (2.68e-3) - & 4.6381e-1 (3.03e-4) - & 4.6034e-1 (1.83e-3) - & \cellcolor[rgb]{ .647,  .647,  .647}{4.6468e-1 (2.22e-4)} \\
		
		CTP7  & 5.6701e-1 (2.23e-3) - & 5.6676e-1 (9.22e-4) - & 5.6637e-1 (4.11e-4) - & \cellcolor[rgb]{ .647,  .647,  .647}{5.6745e-1 (1.69e-4) $\approx$} & 5.6692e-1 (1.86e-4) - & 5.6721e-1 (1.62e-3) \\
	
		CTP8  & 3.4937e-1 (2.45e-2) - & 3.6598e-1 (2.71e-3) - & 3.5213e-1 (3.79e-3) - & 3.6932e-1 (8.68e-4) - & 3.5503e-1 (2.42e-2) - & \cellcolor[rgb]{ .647,  .647,  .647}{3.7069e-1 (4.25e-4)} \\
		\midrule
		
		LIRCMOP1 & 1.3114e-1 (2.17e-2) - & 1.9042e-1 (1.06e-2) - & 1.0593e-1 (3.85e-2) - & 1.4954e-1 (1.82e-2) - & 1.8833e-1 (4.61e-2) - & \cellcolor[rgb]{ .647,  .647,  .647}{2.2304e-1 (6.36e-3)} \\
		
		LIRCMOP2 & 2.5580e-1 (2.95e-2) - & 3.2332e-1 (1.35e-2) - & 2.9229e-1 (3.67e-2) - & 2.9325e-1 (2.05e-2) - & 3.2282e-1 (2.82e-2) - & \cellcolor[rgb]{ .647,  .647,  .647}{3.4702e-1 (3.44e-3)} \\
		
		LIRCMOP3 & 1.1697e-1 (2.61e-2) - & 1.4007e-1 (1.86e-2) - & 9.9083e-2 (2.08e-2) - & 1.2942e-1 (2.47e-2) - & 9.1646e-2 (1.42e-2) - & \cellcolor[rgb]{ .647,  .647,  .647}{1.9947e-1 (4.24e-3)} \\
		
		LIRCMOP4 & 2.1773e-1 (2.72e-2) - & 2.4241e-1 (3.29e-2) - & 1.8974e-1 (4.86e-2) - & 2.3242e-1 (3.14e-2) - & 1.8379e-1 (2.31e-2) - & \cellcolor[rgb]{ .647,  .647,  .647}{3.0693e-1 (3.76e-3)} \\
		
		LIRCMOP5 & 8.9983e-2 (1.05e-1) - & 2.4475e-1 (1.22e-2) - & 2.4215e-1 (1.32e-2) - & \cellcolor[rgb]{ .647,  .647,  .647}{2.8700e-1 (5.09e-3) $\approx$} & 2.6214e-1 (8.89e-2) - & 2.8657e-1 (5.90e-3) \\
		
		LIRCMOP6 & 8.6641e-2 (5.68e-2) - & 1.7269e-1 (1.24e-2) - & 1.4582e-1 (3.86e-2) - & 1.9402e-1 (3.37e-3) $\approx$ & \cellcolor[rgb]{ .647,  .647,  .647}{1.9677e-1 (1.59e-4) +} & 1.8377e-1 (3.56e-2) \\
		
		LIRCMOP7 & 2.8752e-1 (6.87e-3) $\approx$ & 2.6957e-1 (2.12e-2) - & 2.8582e-1 (3.18e-3) - & 2.9114e-1 (4.01e-3) $\approx$ & \cellcolor[rgb]{ .647,  .647,  .647}{2.9389e-1 (1.55e-4) +} & 2.8957e-1 (4.05e-3) \\
		
		LIRCMOP8 & 2.8236e-1 (1.47e-2) - & 2.6950e-1 (1.80e-2) - & 2.8424e-1 (1.40e-2) - & 2.9321e-1 (3.40e-3) $\approx$ & \cellcolor[rgb]{ .647,  .647,  .647}{2.9387e-1 (2.04e-4) $\approx$} & 2.9235e-1 (3.25e-3) \\
		
		LIRCMOP9 & 3.5772e-1 (8.15e-2) - & 5.2821e-1 (2.63e-2) $\approx$ & 4.9955e-1 (2.81e-2) - & \cellcolor[rgb]{ .647,  .647,  .647}{5.5712e-1 (8.54e-3) +} & 4.9582e-1 (5.19e-2) - & 5.3708e-1 (2.00e-2) \\
		
		LIRCMOP10 & 5.1193e-1 (6.34e-2) - & 7.0668e-1 (5.59e-3) + & 6.7264e-1 (2.74e-2) - & 7.0659e-1 (3.77e-4) + & \cellcolor[rgb]{ .647,  .647,  .647}{7.0755e-1 (1.24e-4) +} & 7.0630e-1 (4.01e-4) \\
		
		LIRCMOP11 & 5.3737e-1 (1.32e-1) - & 6.9062e-1 (4.88e-3) - & 6.4145e-1 (1.47e-2) - & 6.9392e-1 (7.61e-5) $\approx$ & 6.1686e-1 (4.29e-2) - & \cellcolor[rgb]{ .647,  .647,  .647}{6.9393e-1 (5.71e-5)} \\
		
		LIRCMOP12 & 5.5161e-1 (4.68e-2) - & 6.1582e-1 (9.12e-3) $\approx$ & 6.0522e-1 (7.22e-3) - & \cellcolor[rgb]{ .647,  .647,  .647}{6.1952e-1 (1.29e-3) $\approx$} & 6.0839e-1 (2.46e-2) - & 6.1811e-1 (3.04e-3) \\
		
		LIRCMOP13 & 4.7950e-1 (1.63e-1) - & 5.3426e-1 (4.14e-3) - & 5.4704e-1 (3.37e-4) - & 5.5421e-1 (1.49e-3) - & 5.1626e-1 (3.56e-3) - & \cellcolor[rgb]{ .647,  .647,  .647}{5.5578e-1 (1.23e-3)} \\
		
		LIRCMOP14 & 4.9121e-1 (1.36e-1) - & 5.3744e-1 (4.86e-3) - & 5.4656e-1 (7.33e-4) - & 5.5357e-1 (1.22e-3) - & 5.2944e-1 (4.28e-3) - & \cellcolor[rgb]{ .647,  .647,  .647}{5.5604e-1 (1.16e-3)} \\
		\midrule
		+/-/$\approx$ & 0/33/3 & 3/31/2 & 3/28/5 & 7/20/9 & 3/32/1 &  \\
		\bottomrule
	\end{tabular}%
	\label{tab:performance-HV}%
\end{table*}%

\begin{figure}[t]\setlength{\abovecaptionskip}{0.cm}
	\setlength{\belowcaptionskip}{-0pt}
	\centering
	\includegraphics[width=3in]{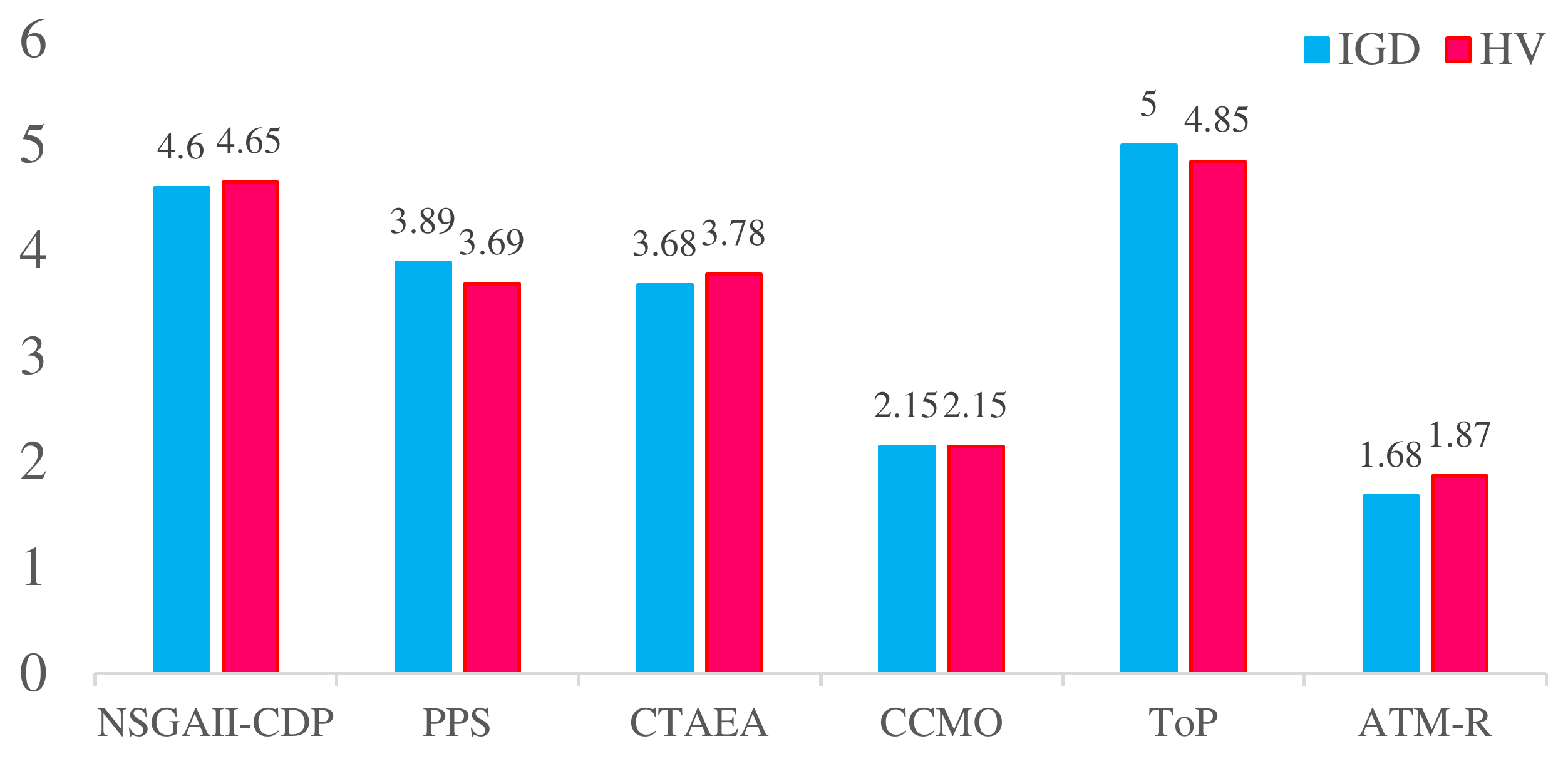}\\
	\caption{Average rankings of six CMOEAs on 36 test functions in terms of the IGD/HV value. A lower ranking value denotes a better performance.}\label{fig:Friedman}
\end{figure}

\subsection{Comparison Results}
First, we compared the performance of ATM-R with that of the other five CMOEAs. The mean IGD values and standard deviations of 36 test functions over 30 independent runs are summarized in Table~\ref{tab:performance-IGD}. The results in terms of the HV value are collected in Table~\ref{tab:performance-HV}.  In each table, ``std" represents the standard deviation of the IGD/HV values over 30 independent runs. ``NaN" denotes that a CMOEA cannot find a feasible solution of a test function over all 30 independent runs. For a given test function, ATM-R was compared with each competitor by the Friedman test with Bonferroni correction at a significance level of 0.05. For convenience,
``+", ``-", and ``$\approx$" are used to represent that a competitor is better than, worse than, and similar to ATM-R, respectively. In addition, for each test function, the best result among the six CMOEAs is highlighted in gray. To visualize the results, we plotted the CPFs obtained by the six CMOEAs in a typical run on three representative CMOPs in Figs. \ref{fig:MW13}-\ref{fig:LIRCMOP14}. A typical run denotes the one producing
the median IGD value among all runs. 

\subsubsection{General Performance} 
\begin{figure*}[ht]\setlength{\abovecaptionskip}{-0cm}\setlength{\belowcaptionskip}{-0cm}
	\begin{center}
		
		\subfigure{\includegraphics[width=0.32\columnwidth]{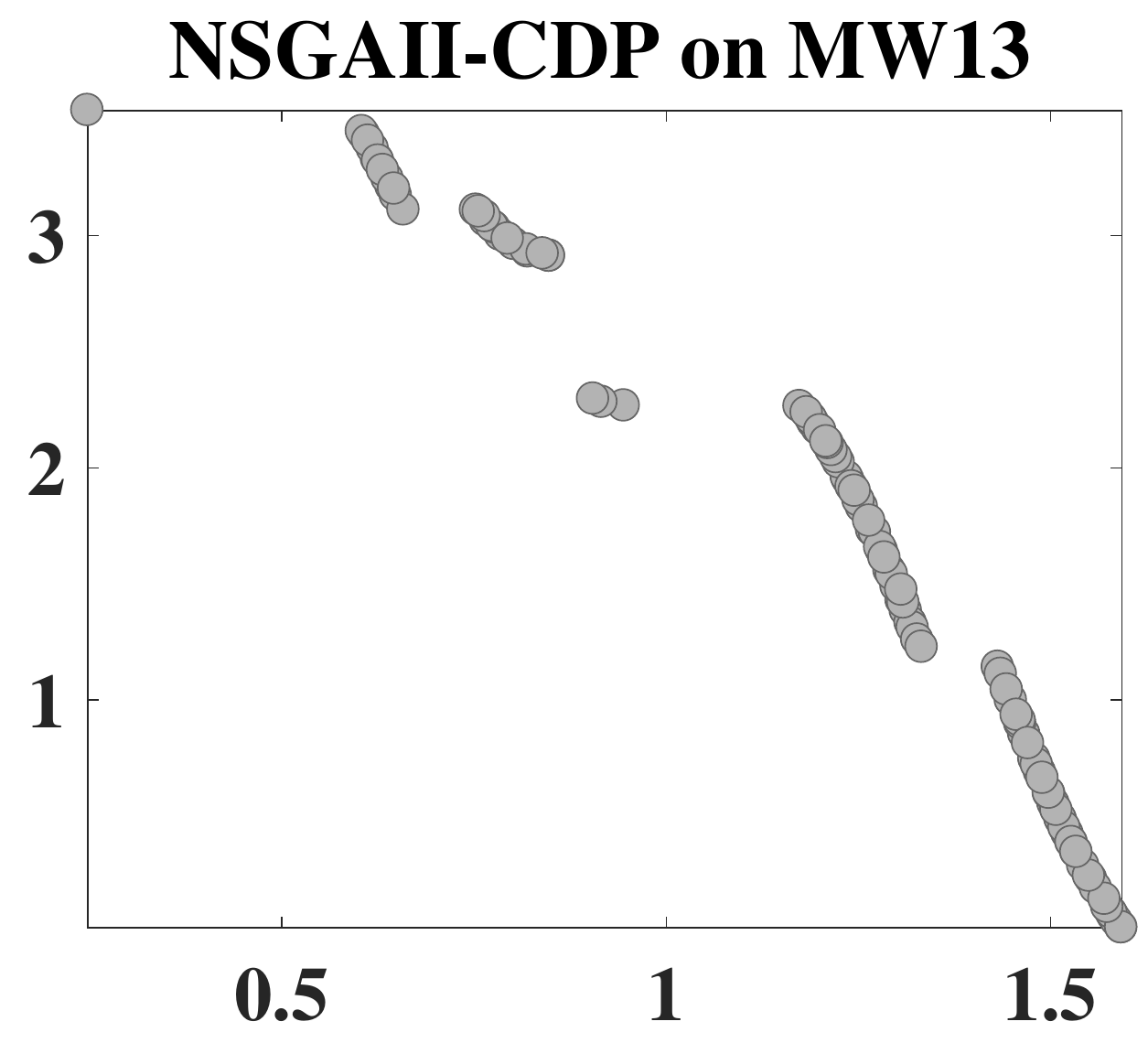}}
		\subfigure{\includegraphics[width=0.325\columnwidth]{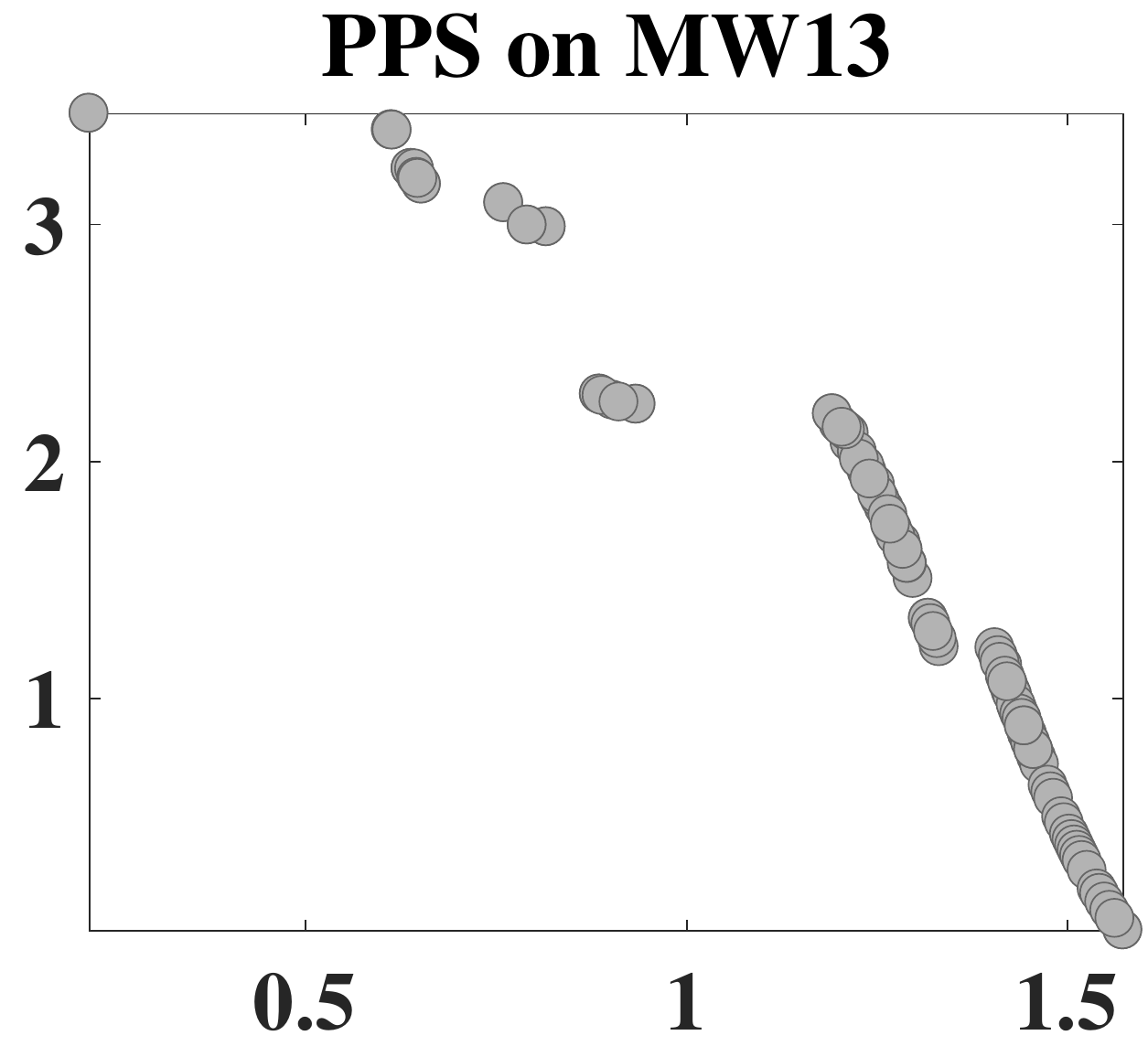}}
		\subfigure{\includegraphics[width=0.33\columnwidth]{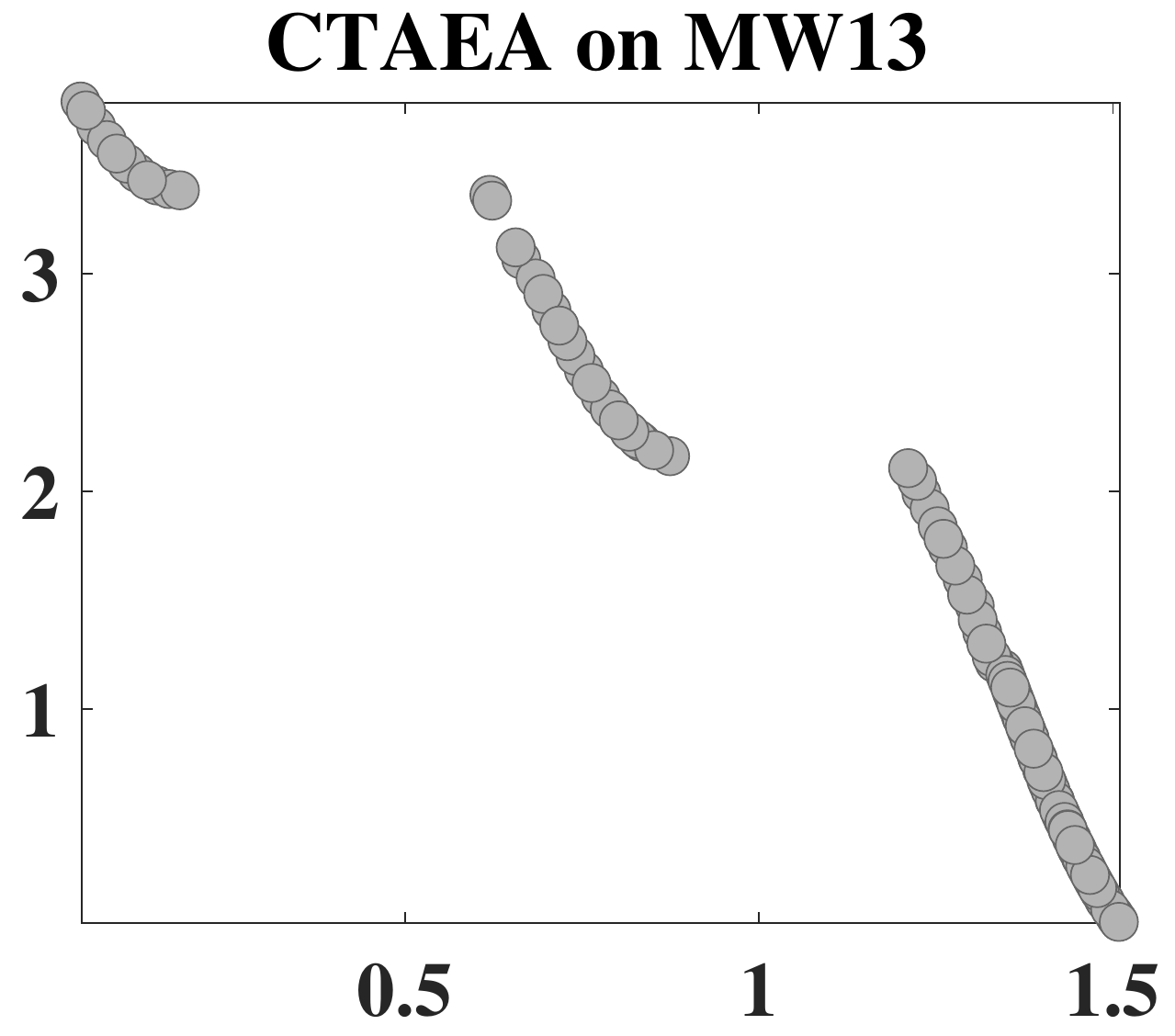}}
		\subfigure{\includegraphics[width=0.325\columnwidth]{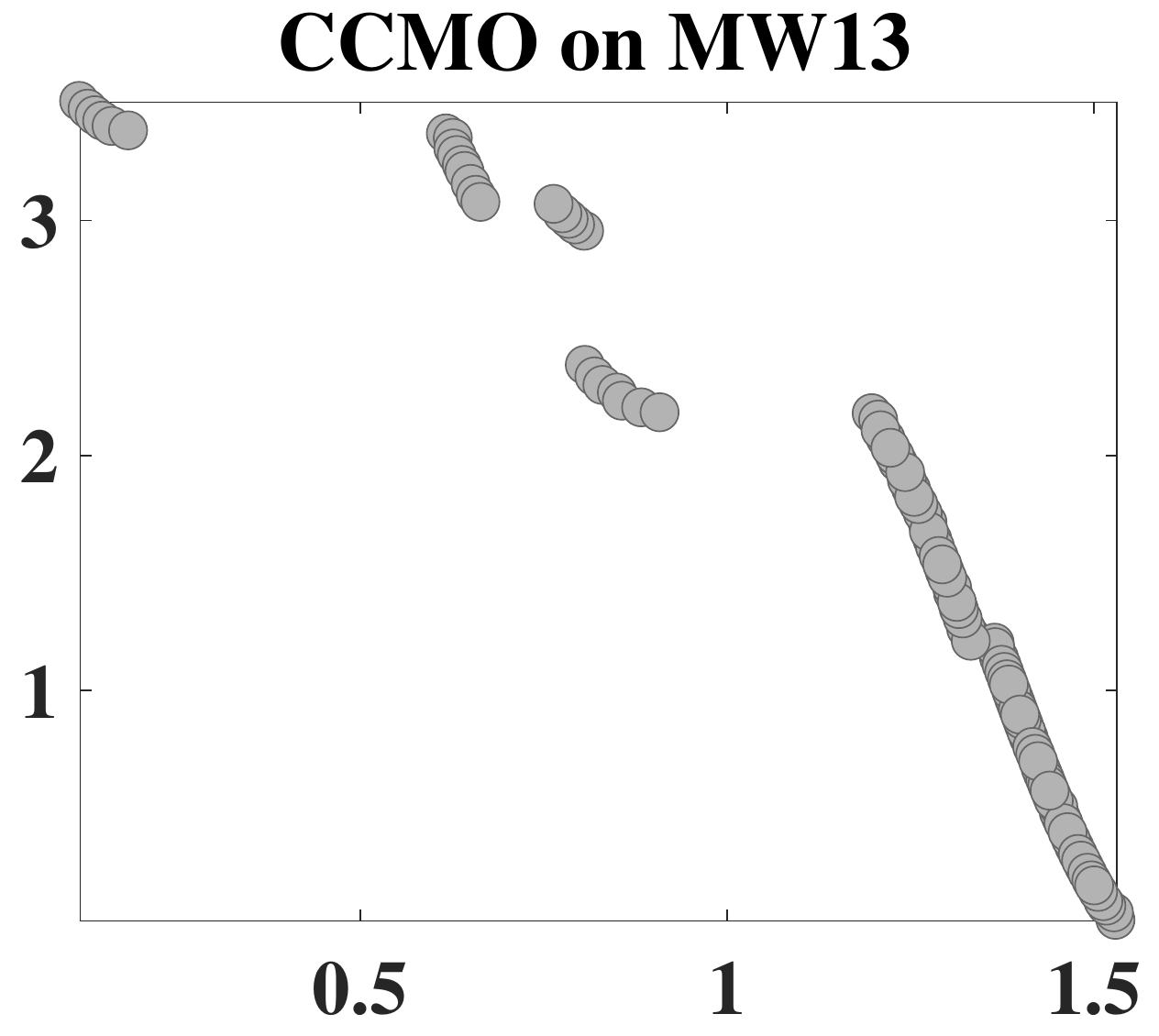}}
		\subfigure{\includegraphics[width=0.33\columnwidth]{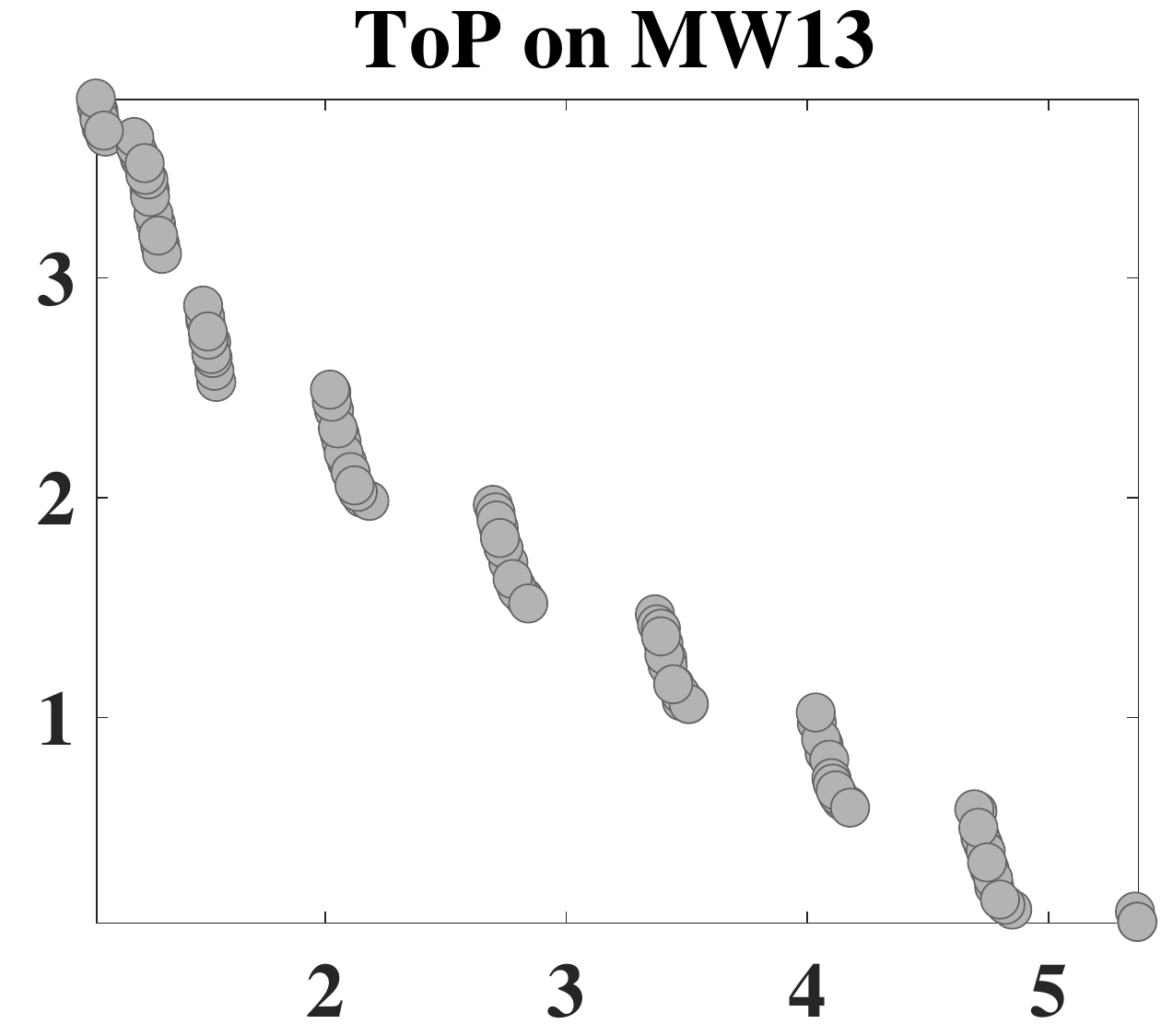}}
		\subfigure{\includegraphics[width=0.33\columnwidth]{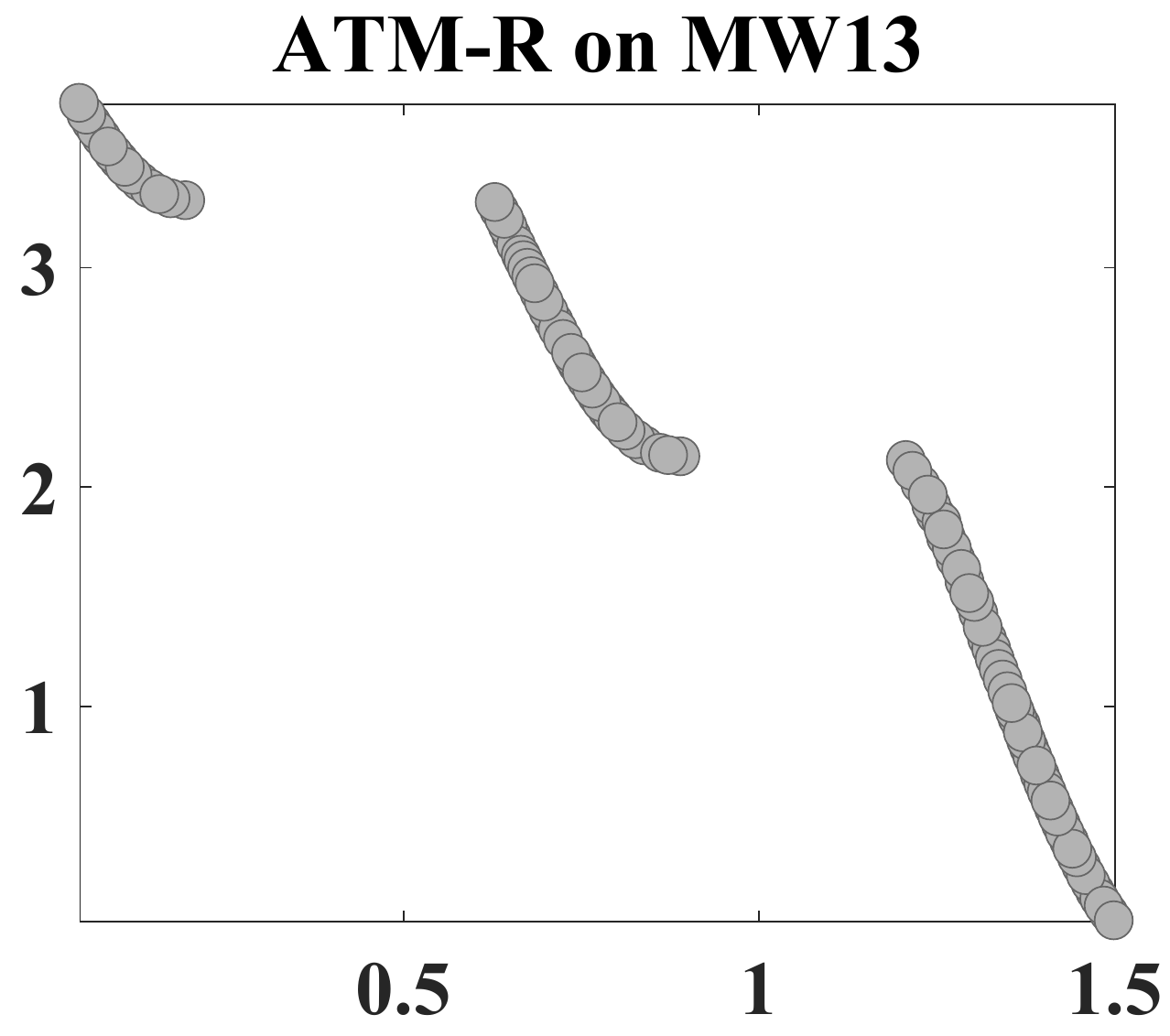}}
		\caption{The constrained Pareto front with median value among 30 runs obtained by NSGAII-CDP, PPS, CTAEA, CCMO, ToP, and ATM-R on MW13.}\label{fig:MW13}
	\end{center}
\end{figure*}

\begin{figure*}[ht]\setlength{\abovecaptionskip}{-0cm}\setlength{\belowcaptionskip}{-0cm}
	\begin{center}
		
		\subfigure{\includegraphics[width=0.33\columnwidth]{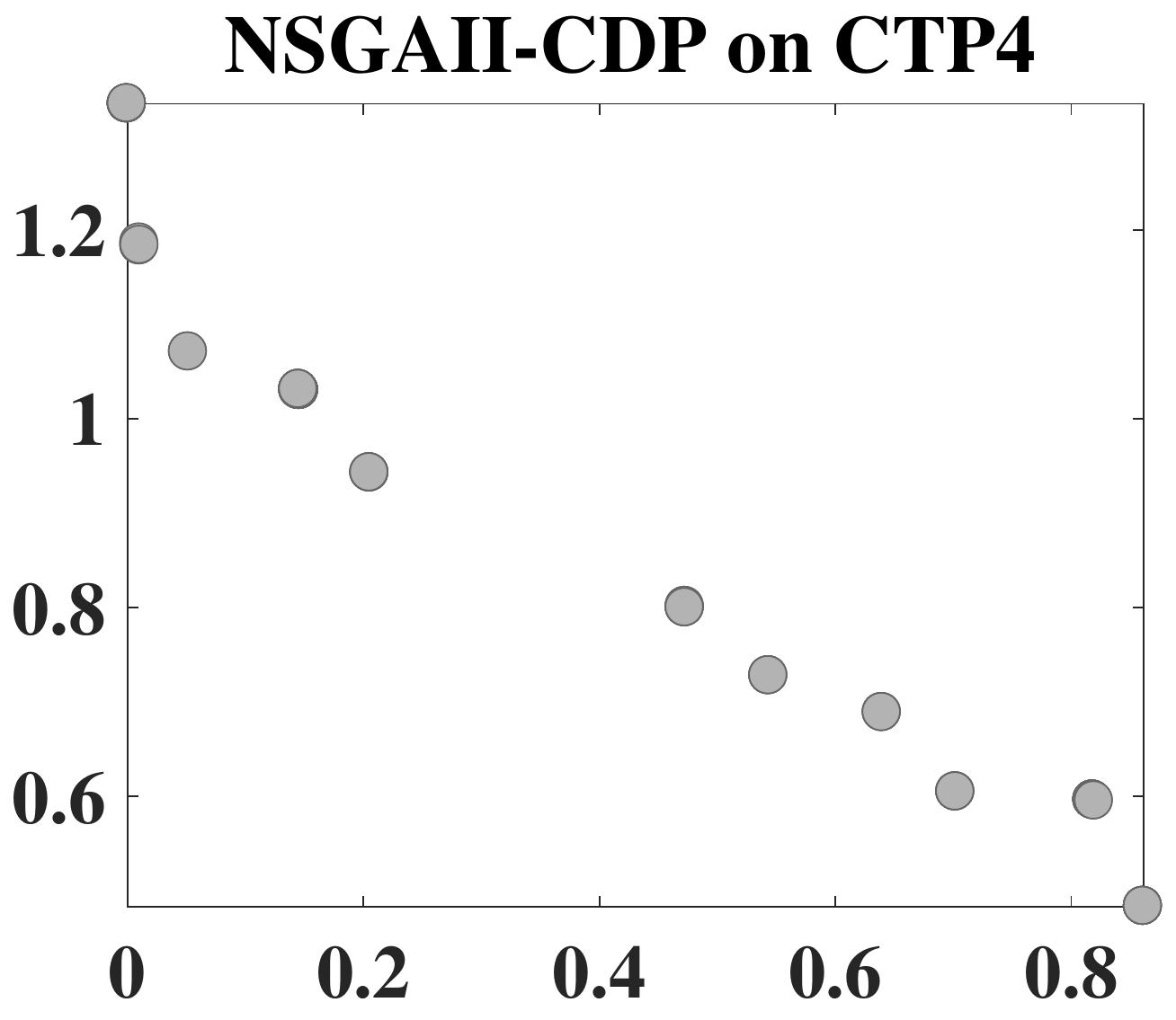}}
		\subfigure{\includegraphics[width=0.33\columnwidth]{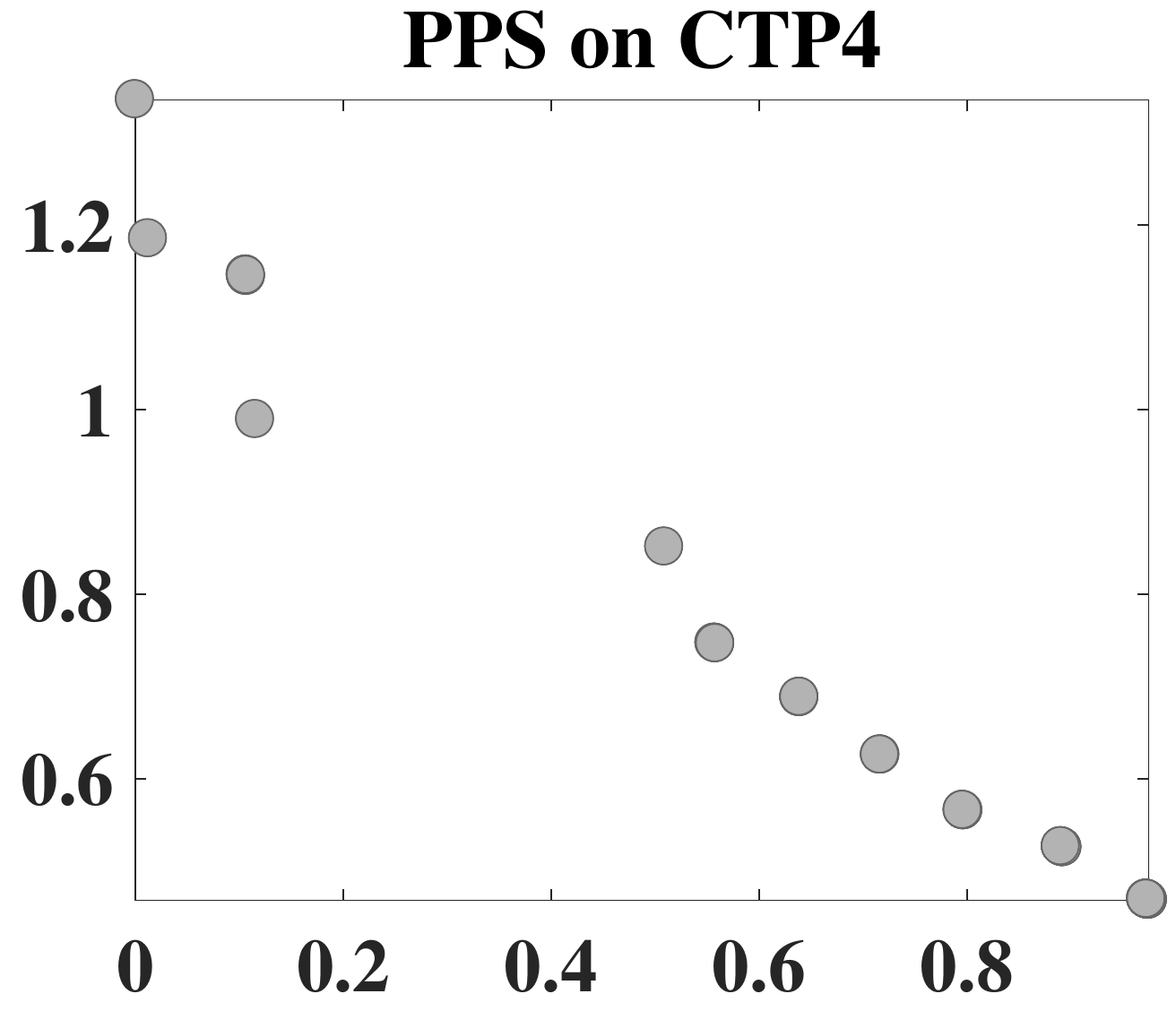}}
		\subfigure{\includegraphics[width=0.33\columnwidth]{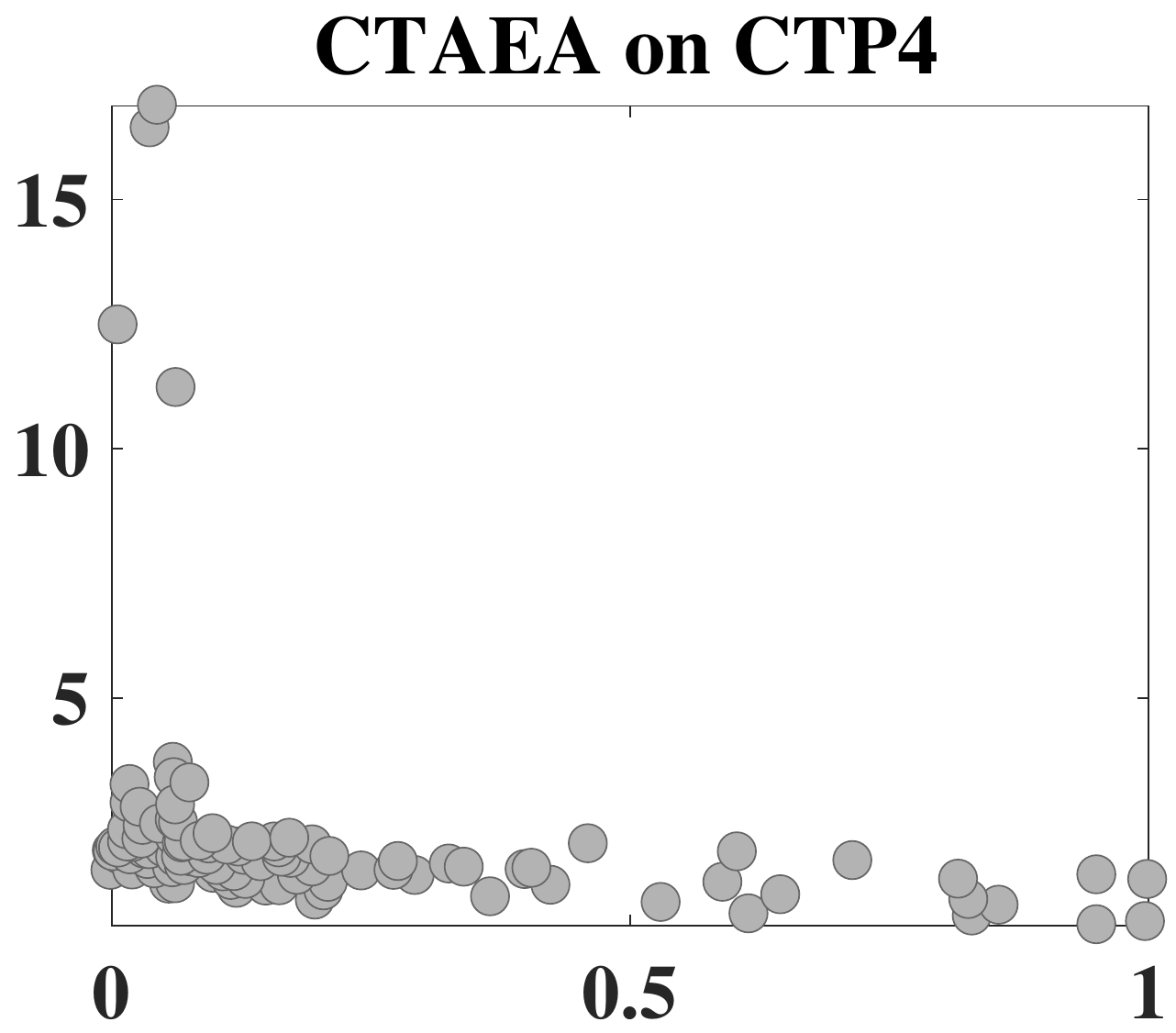}}
		\subfigure{\includegraphics[width=0.33\columnwidth]{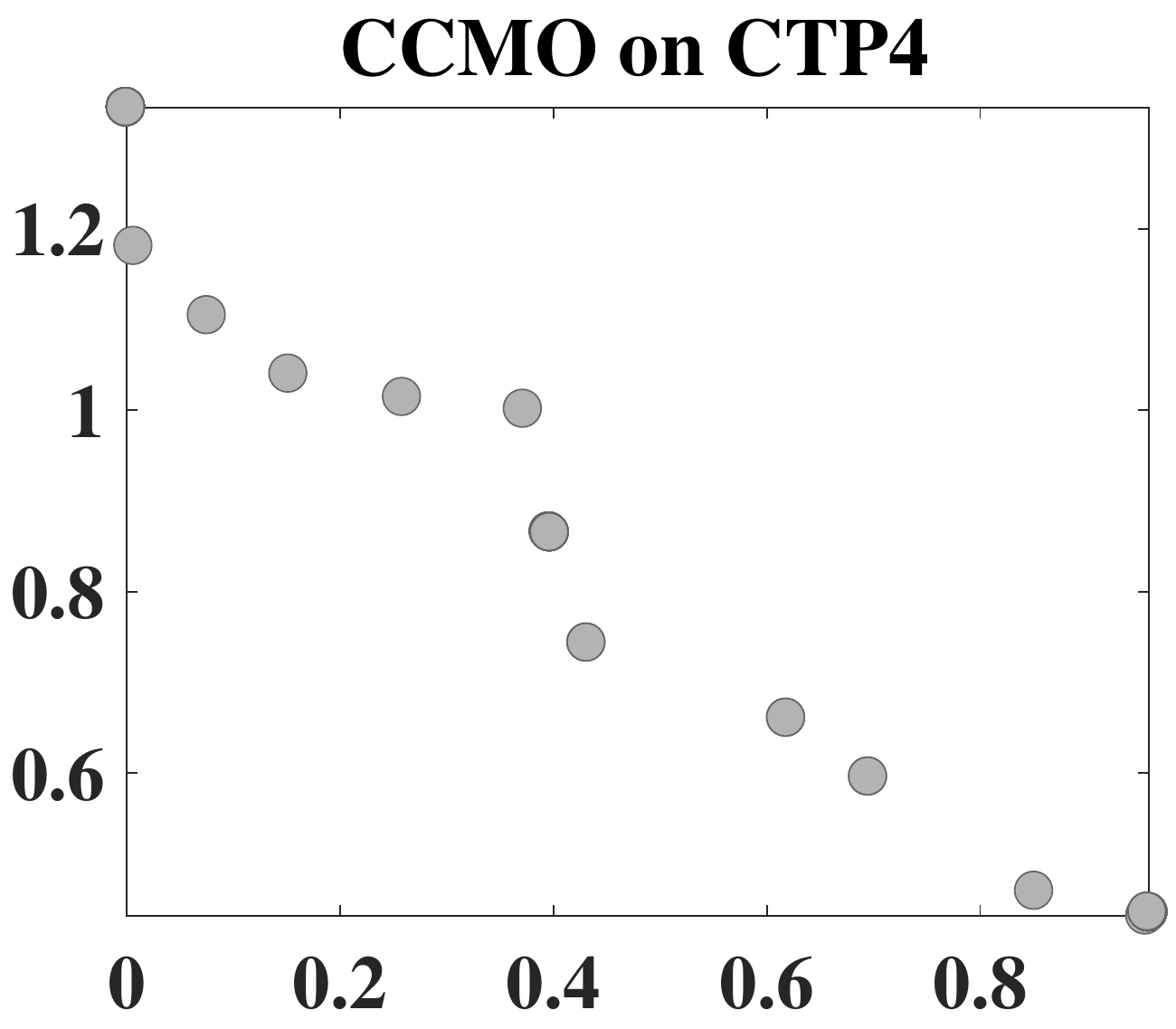}}
		\subfigure{\includegraphics[width=0.33\columnwidth]{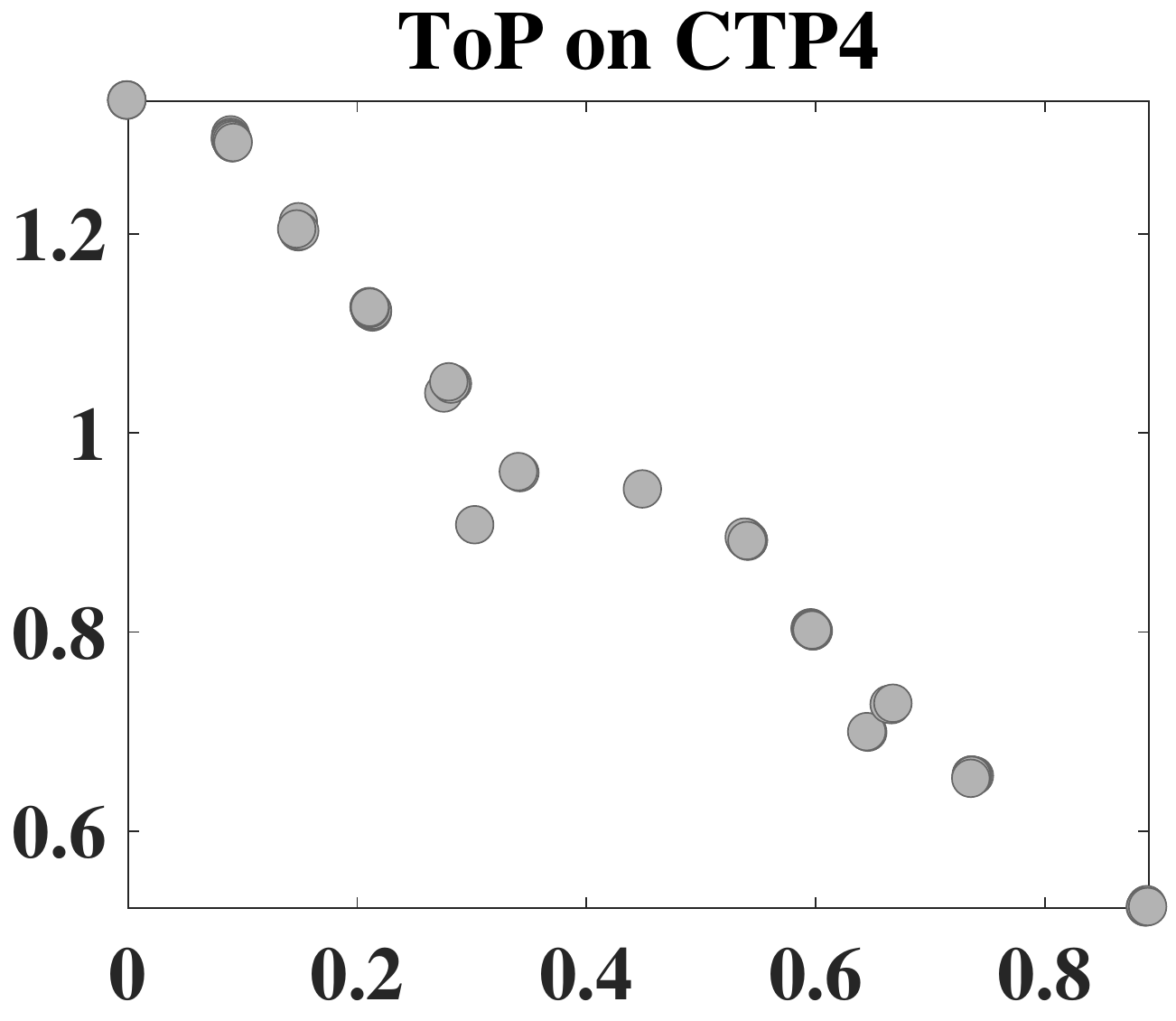}}
		\subfigure{\includegraphics[width=0.33\columnwidth]{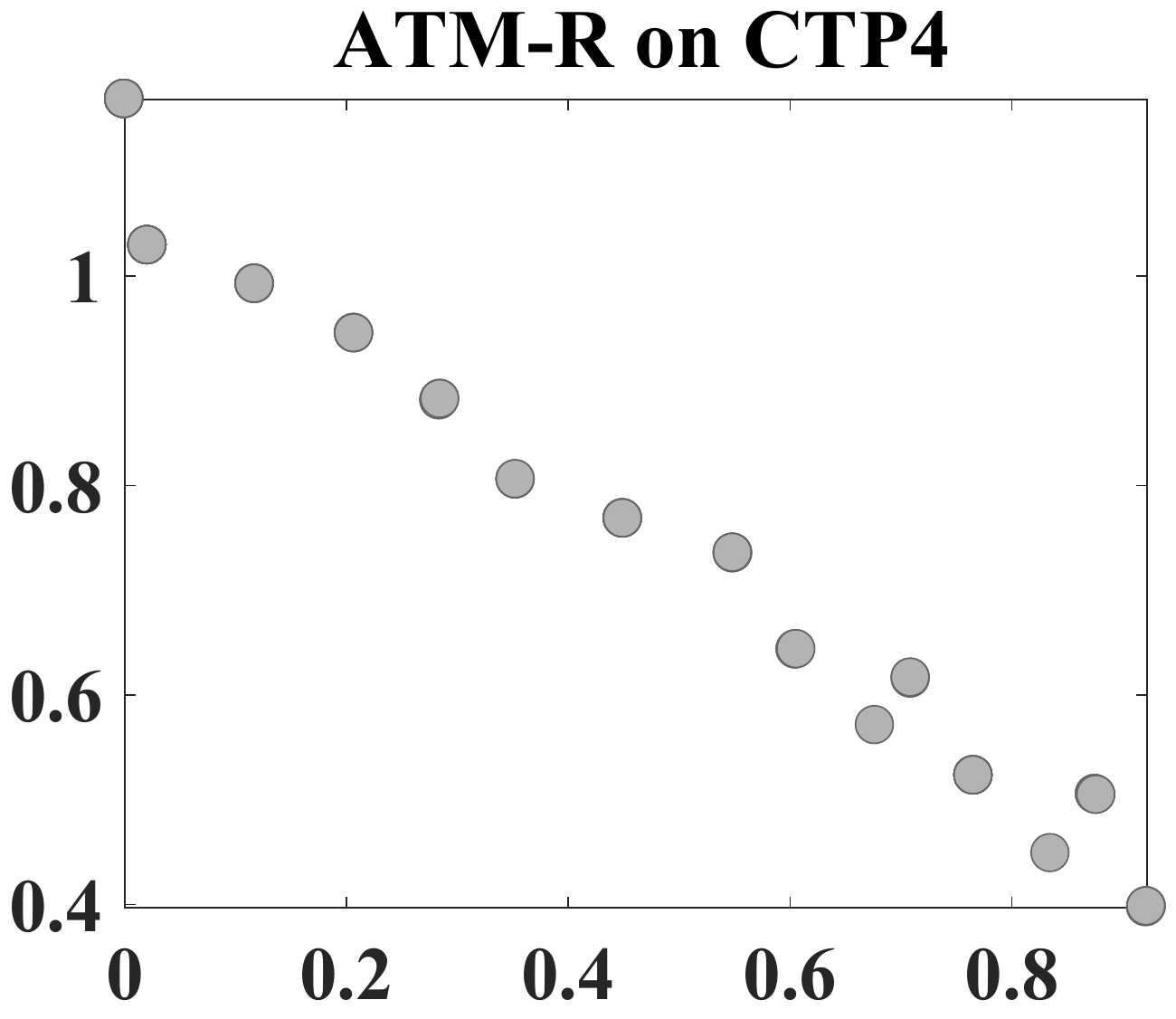}}
		\caption{The constrained Pareto front with median value among 30 runs obtained by NSGAII-CDP, PPS, CTAEA, CCMO, ToP, and ATM-R on CTP4.}\label{fig:CTP4}
	\end{center}
\end{figure*}

\begin{figure*}[ht]\setlength{\abovecaptionskip}{-0cm}\setlength{\belowcaptionskip}{-0cm}
	\begin{center}
		
		\subfigure{\includegraphics[width=0.33\columnwidth]{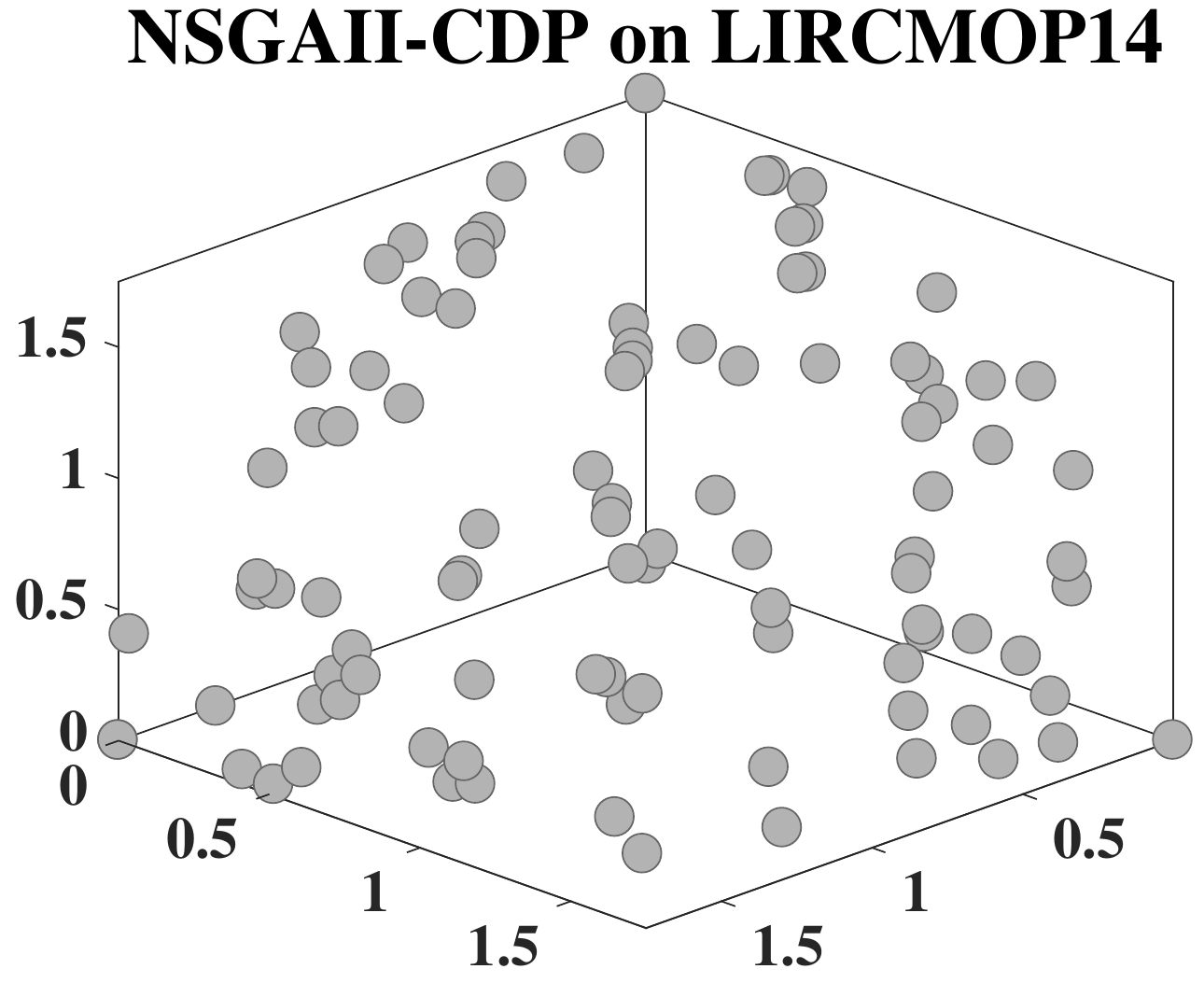}}
		\subfigure{\includegraphics[width=0.335\columnwidth]{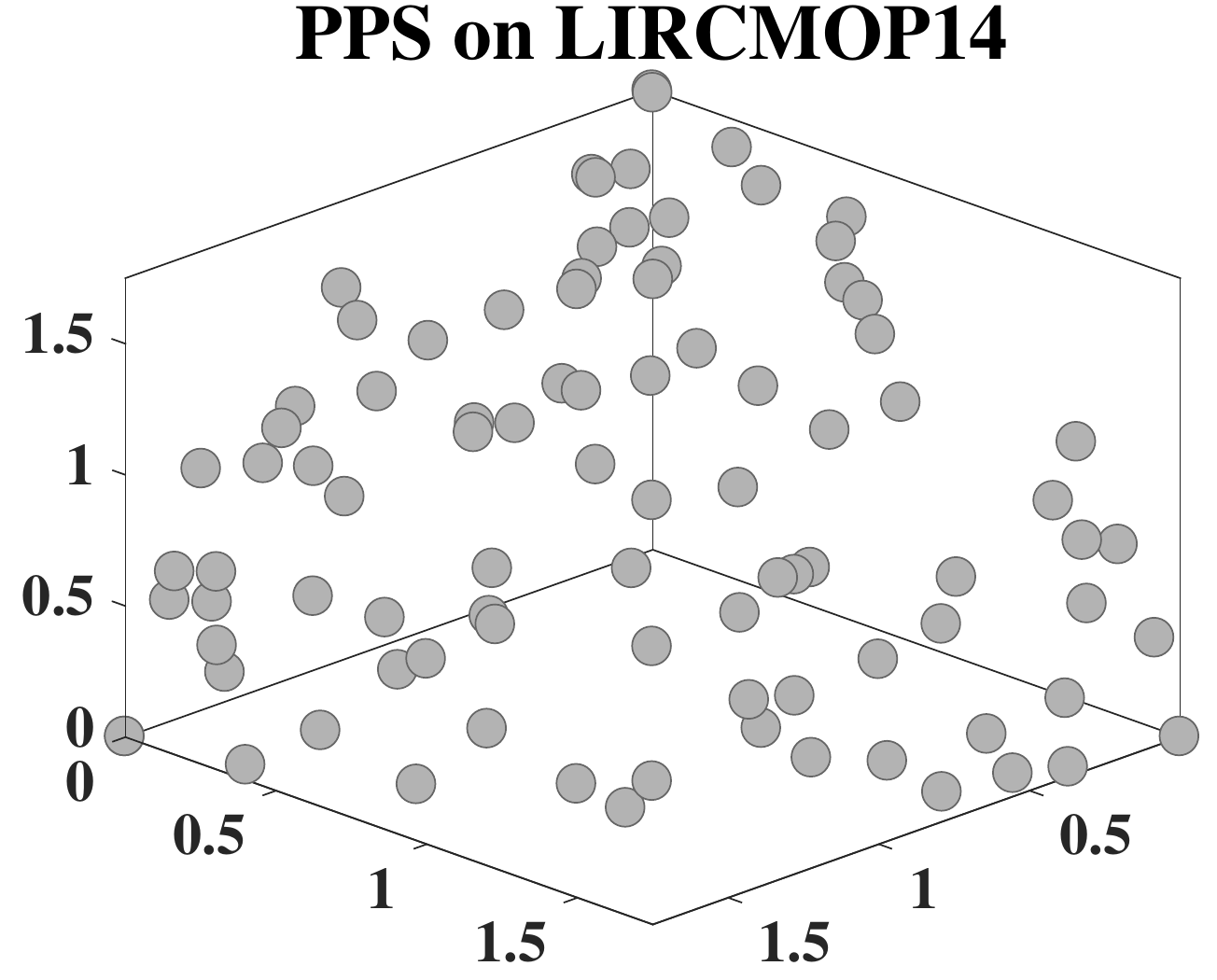}}
		\subfigure{\includegraphics[width=0.33\columnwidth]{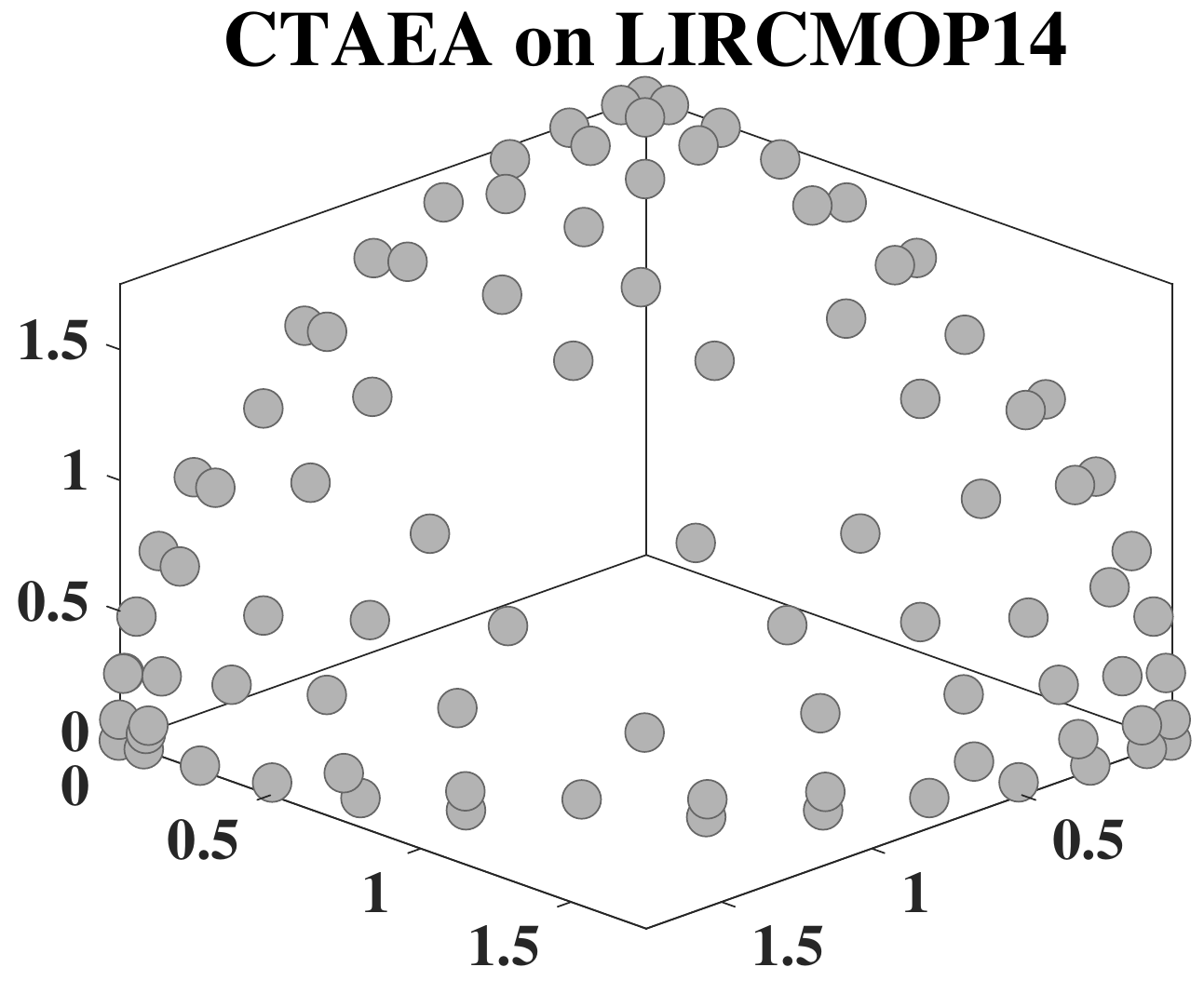}}
		\subfigure{\includegraphics[width=0.33\columnwidth]{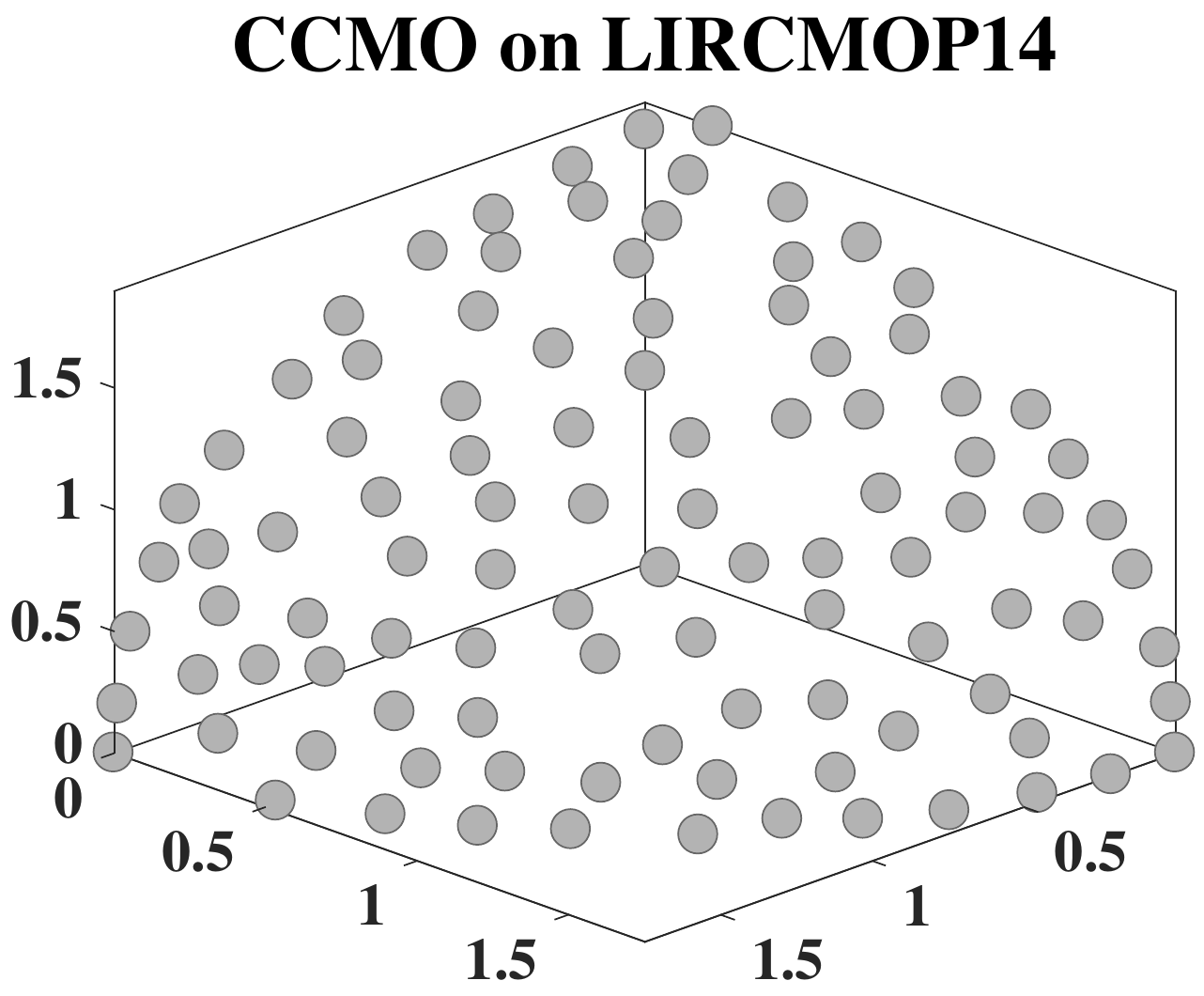}}
		\subfigure{\includegraphics[width=0.33\columnwidth]{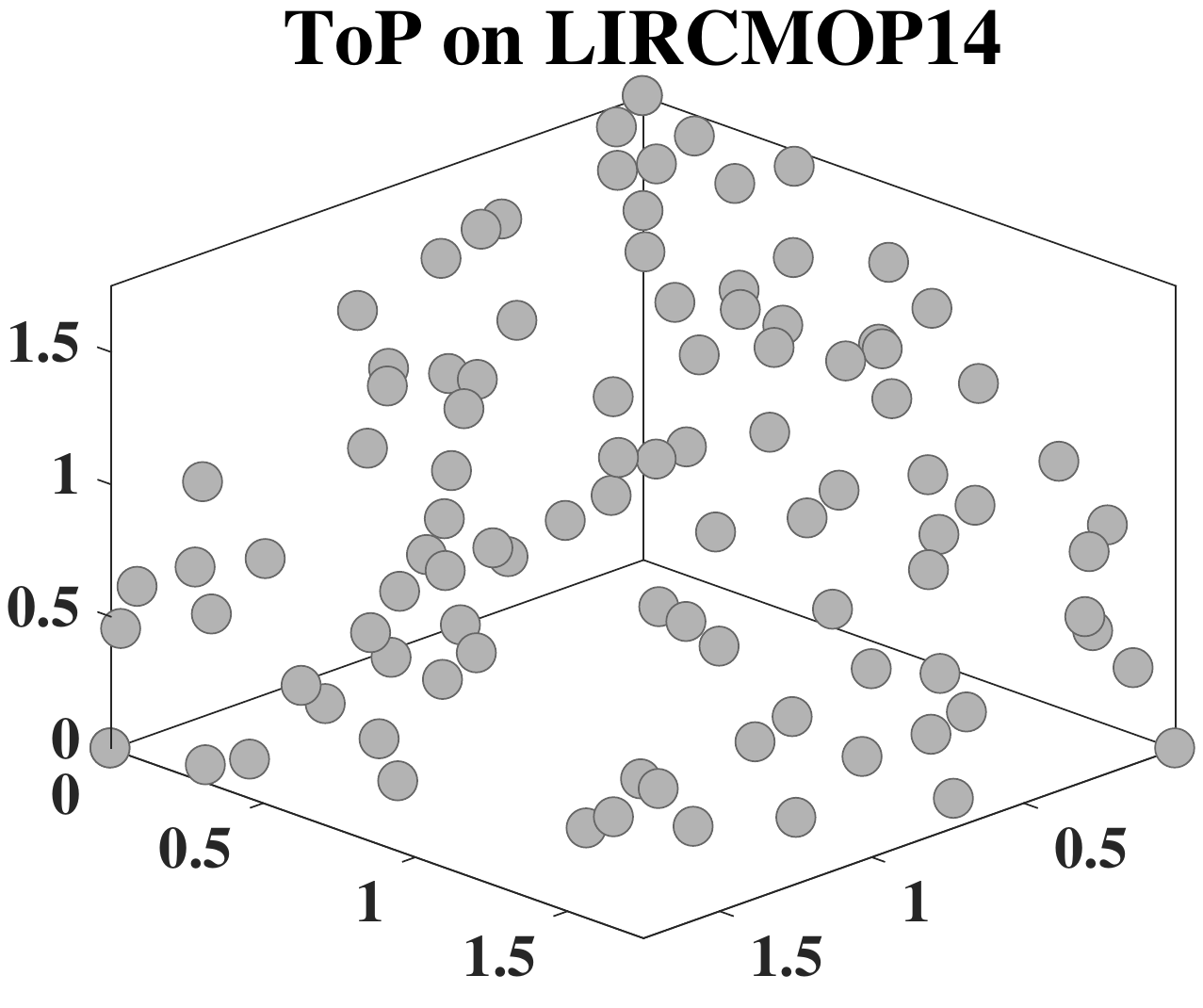}}
		\subfigure{\includegraphics[width=0.33\columnwidth]{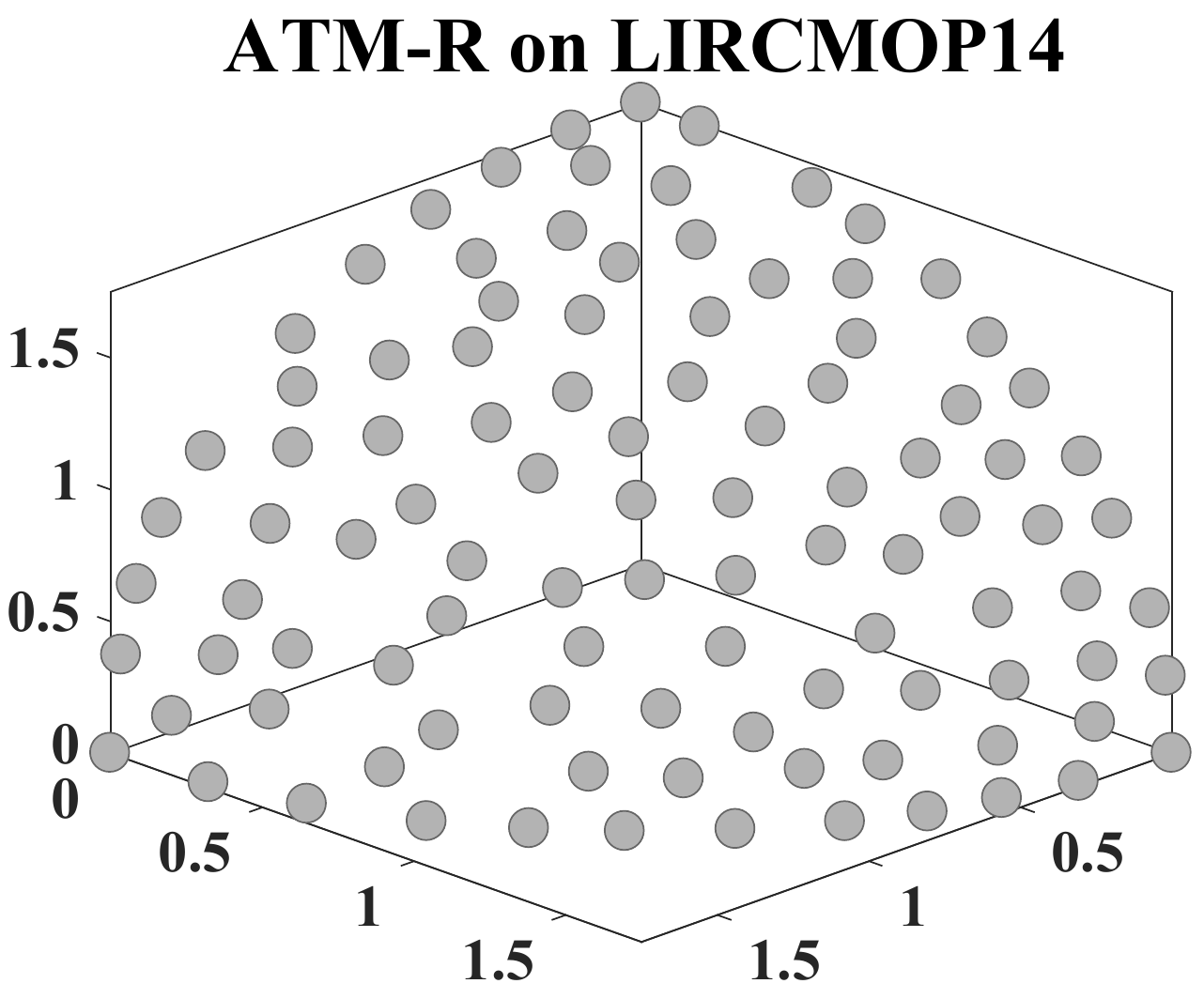}}
		\caption{The constrained Pareto front with median value among 30 runs obtained by NSGAII-CDP, PPS, CTAEA, CCMO, ToP, and ATM-R on LIRCMOP14.}\label{fig:LIRCMOP14}
	\end{center}
\end{figure*}

In general, as shown in Table~\ref{tab:performance-IGD} and Table~\ref{tab:performance-HV}, ATM-R obtained the best results of most of the test functions in terms of both the IGD and the HV values. Additionally, it performed significantly better than the other five competitors on most of the test functions. The multi-problem Friedman's test~\cite{alcala2009keel} was implemented to compare these six CMOEAs simultaneously. As shown in Fig.~\ref{fig:Friedman}, ATM-R achieved the lowest ranking value among six CMOEAs. Furthermore, the results in Figs. \ref{fig:MW13}-\ref{fig:LIRCMOP14} show that ATM-R can obtain a set of well-converged and well-distributed solutions. A more detailed discussion on different test suites is given next.


\subsubsection{Performance on MW Test Suite} 
In terms of the IGD value, ATM-R performed better than NSGAII-CDP, PPS, CTAEA, CCMO, and ToP on 14, 13, six, six, and 14 test functions, respectively. Inversely, these peer CMOEAs were better than ATM-R on zero, one, two, three, and zero test functions, respectively. ATM-R obtained the best results of four test functions on which it performed better than the other five competitors. CCMO obtained the best results of four test functions, on one of which it performed similarly to ATM-R. Although CTAEA obtained the best results of six test functions, it performed similarly to ATM-R on five of these test functions. 

In terms of the HV value, ATM-R performed better than NSGAII-CDP, PPS, CTAEA, CCMO, and ToP on 12, 12, six, eight, and 14 test functions, respectively. On the contrary, these peer CMOEAs revealed better results than ATM-R on zero, two, three, five, and zero test functions, respectively. ATM-R obtained the best results of three test functions on which it performed better than the other five competitors. CCMO obtained the best results of four test functions. Although CTAEA obtained the best results of seven test functions, it performed similarly to ATM-R on four of these test functions.      

Furthermore, as shown in Fig~\ref{fig:MW13}, ATM-R obtained a set of well-converged and well-distributed feasible solutions that is close to the CPF. However, ToP failed to converge to the CPF. NSGAII-CDP, PPS, CTAEA, and CCMO lost some parts of the CPF. The results reflect that ATM-R performs better than the other five competitors on the MW test suite.   

\subsubsection{Performance on CTP Test Suite}
For the CTP test suite, ATM-R performed better than the other five competitors on most of the test functions in terms of both the IGD and HV values. Additionally, it obtained the best IGD/HV values on most of the test functions.  

For CTP1, some parts of the CPF come from the unconstrained Pareto front. For CTP6, the objective space has infeasible holes of differing widths toward the Pareto-optimal regions. For CTP2-CTP5, CTP7, and CTP8, the CPFs are divided into several disconnected segments. To solve these problems effectively, infeasibility information should be used carefully. Thus, NSGAII-CDP and ToP, which only consider constraints in the infeasible phase, performed worse than ATM-R. As stated in~\cite{liu2021handling}, due to the complex (i.e., disconnected and discrete) characteristics  of the CPFs, the CMOEAs using reference points or vectors would have inferior performance. Therefore, CTAEA performed worse than ATM-R.  PPS puts emphasis on objectives in the early stage, while CCMO adopts a specific population to make use of infeasibility information.  Compared with the MW test suite, the test functions in the CTP test suite have larger feasibility ratios. Thus, too much infeasibility information would impair the performance of a CMOEA. This may be why PPS and CCMO performed worse than ATM-R.

Furthermore, as shown in Fig.~\ref{fig:CTP4}, ATM-R can converge to the CPF of CTP4 more quickly than the other five competitors. Additionally, it can cover more parts of the CPF than the other five competitors. The results reflect that ATM-R performs better than the other five competitors on the CTP test suite. 

\subsubsection{Performance on LIRCMOP Test Suite}
For the LIRCMOP test suite, ATM-R obtained the best results on half of the test functions in terms of both the IGD and HV values. Similar to the CTP test suite, the test functions in the LIRCMOP test suite have infeasible holes in the objective space and the CPFs of some test functions are disconnected.
To solve these test functions effectively, infeasibility information should be used carefully. NSGAII-CDP and ToP performed worse than ATM-R on most of the test functions because they ignore the infeasibility information to a great extent. ToP performed better than ATM-R on four and three test functions in terms of the IGD and HV values, respectively. This is attributed to the powerful genetic operator (i.e., differential evolution) used in ToP. 

Among the infeasibility-assisted CMOEAs, PPS motivates the population toward the unconstrained Pareto front in the early stage. In CTAEA and CCMO, an additional population is employed to approach the unconstrained Pareto front. Thus, these three CMOEAs will fail to solve a CMOP (i.e., LIRCMOP1-LIRCMOP4) in which the CPF is far away from the unconstrained Pareto front. Regarding LIRCMOP5 and LIRCMOP6, the CPFs are the same as the unconstrained Pareto fronts. 
Regarding LIRCMOP7 and LIRCMOP8, the CPFs are near the unconstrained Pareto fronts. For these four test functions, an infeasibility-assisted CMOEA can approach the CPF easily; thus, uniformity is the key factor affecting its performance. ATM-R and CCMO performed better than PPS and CTAEA on these test functions because they can preserve diversity more effectively. For LIRCMOP9-LIRCMOP12, the CPFs are divided into several disconnected segments. To solve these test functions effectively, diversity should be maintained carefully. Due to the weak cooperation of two populations, CCMO can maintain diversity effectively during the evolutionary process. Thus, it performed better than ATM-R on these four test functions. For the two three-objective test functions (i.e., LIRCMOP13-LIRCMOP14), ATM-R performed better than the other five competitors in terms of both the IGD and HV values. It implies that ATM-R can achieve a better tradeoff among feasibility, diversity, and convergence for three-objective CMOPs.
       
Furthermore, as shown in Fig.~\ref{fig:LIRCMOP14}, all six CMOEAs can converge to the constrained Pareto front successfully. ATM-R performed better than the other five competitors in terms of the uniformity since it can achieve a better tradeoff among feasibility, diversity, and convergence for three-objective CMOPs. The results reflect that ATM-R performs better than the other five competitors on the LIRCMOP test suite. 

In summary, the extensive experiments on 36 test functions with various challenging characteristics demonstrate that ATM-R is able to solve complex CMOPs successfully.

\section{Further Analyses}

\subsection{Advantages of ATM-R over ATM}

As discussed in {\it Remark 1}, it is not effective to extend ATM~\cite{wang2008adaptive} to solve CMOPs straightforwardly. In this subsection, the advantages of ATM-R over ATM were demonstrated through experiments. The comparison results on 36 test functions in terms of the IGD and the HV values are summarized in Table~\ref{tbl:comparison-ATM}, where ``+", ``-", and ``$\approx$" denote that ATM performs better than, worse than, and similarly to ATM-R in terms of the IGD/HV value, respectively. As shown in Table~\ref{tbl:comparison-ATM}, ATM-R performed better than ATM on these three test suites in terms of both the IGD and the HV values. Specifically, in terms of the IGD value, ATM-R was better than ATM on 9, 5, and 12 test functions of the MW, the CTP, and the LIRCMOP test suites, respectively. In contrast, ATM was better than ATM-R on no more than three test functions of these test suites. With regard to the HV value, ATM-R performed better than ATM on 8, 6, and 12 test functions, respectively. Inversely, ATM outperformed ATM-R on no more than 4 test functions of these test suites. In summary, the experimental results on these test functions with various characteristics demonstrate that ATM-R has an edge over ATM.

\begin{table}[t]
	\centering
	\caption{Results of ATM vs ATM-R on 36 test functions.}
	\begin{tabular}{ccc}
		\toprule
		Test Functions &\tabincell{c}{IGD\\ +/-/$\approx$}  & \tabincell{c}{HV\\ +/-/$\approx$} \\
		\midrule
		MW1-MW14   & 2/9/3   & 4/8/2 \\
		CTP1-CTP8  & 3/5/0   & 2/6/0 \\
		LIRCMOP1-LIRCMOP14  & 1/12/1    & 1/12/1  \\
		\bottomrule
	\end{tabular}%
	\label{tbl:comparison-ATM}%
\end{table}%

\subsection{Effectiveness of the Infeasible Phase}

To validate the effectiveness of the update mechanism in the infeasible phase, we implemented three variants (denoted as ATM-RICDP, ATM-RIobj, and ATM-RIdiv) by using different update mechanisms in this phase. Specifically, in ATM-RICDP, the CDP is used for solution selection. In ATM-RIobj, the solutions are selected based on Pareto dominance regardless of constraints. In ATM-RIdiv, the diversity is quantified and used to select promising solutions. By comparing ATM-R with each of ATM-RICDP, ATM-RIobj, and ATM-RIdiv, the effectiveness of the update mechanism in the infeasible phase can be validated. The comparison results on 36 test functions mentioned above are summarized in Table~\ref{tbl:comparison-first},  where ``+", ``-", and ``$\approx$" denote that a competitor performs better than, worse than, and similarly to ATM-R in terms of the IGD/HV value, respectively.

\begin{table}[t]
	\centering
	\caption{Results of ATM-RICDP vs ATM-R, ATM-RIobj vs ATM-R, and ATM-RIdiv vs ATM-R on 36 test functions.}
	\begin{tabular}{ccc}
		\toprule
		Algorithms &\tabincell{c}{IGD\\ +/-/$\approx$}  & \tabincell{c}{HV\\ +/-/$\approx$} \\
		\midrule
		ATM-RICDP vs ATM-R  & 2/19/15  & 2/23/11 \\
		ATM-RIobj vs ATM-R   & 5/11/20	& 7/11/18 \\
		ATM-RIdiv vs ATM-R   & 8/14/14   & 	10/14/12 \\
		\bottomrule
	\end{tabular}%
	\label{tbl:comparison-first}%
\end{table}%

As shown in Table~\ref{tbl:comparison-first}, ATM-R performed better than ATM-RICDP, ATM-RIobj, and ATM-RIdiv in terms of both the IGD and the HV values. Specifically, with regard to the IGD value, ATM-R was better than ATM-RICDP, ATM-RIobj, and ATM-RIdiv on 19, 11, and 14 test functions, respectively. Inversely, ATM-RICDP, ATM-RIobj, and ATM-RIdiv outperformed ATM-R on 2, 5, and 8 test functions, respectively. In terms of the HV value, ATM-R performed better than ATM-RICDP, ATM-RIobj, and ATM-RIdiv on 23, 11, and 14 test functions, respectively. In contrast, ATM-RICDP, ATM-RIobj, and ATM-RIdiv outperformed ATM-R on 2, 7, and 4 test functions, respectively. The experimental results show that the update mechanism in the infeasible phase is critical to ATM-R.

\subsection{Effectiveness of the Semi-feasible Phase}

To validate the effectiveness of the update mechanism in the semi-feasible phase,  we implemented three variants (i.e., ATM-RSCDP, ATM-RSobj, and ATM-RSdiv) by using different update mechanisms in this phase. Specifically, in ATM-RSCDP, ATM-RSobj, and ATM-RSdiv, the CDP, the Pareto dominance, and the diversity are used for solution selection, respectively. By comparing ATM-R with each of ATM-RICDP, ATM-RIobj, and ATM-RIdiv, the effectiveness of the update mechanism in the semi-feasible phase can be validated. Specifically, the comparison results on 36 test functions mentioned above are summarized in Table~\ref{tbl:comparison-second},  where ``+", ``-", and ``$\approx$" denote that a competitor performs better than, worse than, and similarly to ATM-R in terms of the IGD/HV value, respectively.

As shown in Table~\ref{tbl:comparison-second}, ATM-R performed better than ATM-RSCDP, ATM-RSobj, and ATM-RSdiv in terms of both the IGD and the HV values. With regard to the IGD value, ATM-R was better than ATM-RSCDP, ATM-RSobj, and ATM-RSdiv on 24, 32, and 36 test functions, respectively. In contrast, ATM-RSCDP, ATM-RSobj, and ATM-RSdiv outperformed ATM-R on no more than three test functions. In terms of the HV value, ATM-R performed better than ATM-RSCDP, ATM-RSobj, and ATM-RSdiv on 23, 31, and 36 test functions, respectively. Inversely, ATM-RSCDP, ATM-RSobj, and ATM-RSdiv outperformed ATM-R on no more than five test functions. The experimental results show that the update mechanism in the semi-feasible phase is critical to ATM-R.

\begin{table}[t]
	\centering
	\caption{Results of ATM-RSCDP vs ATM-R, ATM-RSobj vs ATM-R, and ATM-RSdiv vs ATM-R on 36 test functions.}
	\begin{tabular}{ccc}
		\toprule
		Algorithms &\tabincell{c}{IGD\\ +/-/$\approx$}  & \tabincell{c}{HV\\ +/-/$\approx$} \\
		\midrule
		ATM-RSCDP vs ATM-R  & 3/24/9   & 5/23/8 \\
		ATM-RSobj vs ATM-R   & 3/32/1   & 2/31/3 \\
		ATM-RSdiv vs ATM-R   & 0/36/0    & 0/36/0  \\
		\bottomrule
	\end{tabular}%
	\label{tbl:comparison-second}%
\end{table}%

\subsection{Effectiveness of the Multiphase Mating Selection Strategy}
In order to verify the effectiveness of the multiphase mating selection strategy, we implemented three variants (i.e., ATM-RMCDP, ATM-RMobj, and ATM-RMdiv) by using different selection methods to select mating solutions. Specifically, in ATM-RMCDP, ATM-RMobj, and ATM-RMdiv, the CDP, the Pareto dominance, and the diversity are used for solution selection, respectively. By comparing ATM-R with each of ATM-RMCDP, ATM-RMobj, and ATM-RMdiv, the effectiveness of the multiphase mating selection strategy can be demonstrated. Specifically, the comparison results on 36 test functions mentioned above are summarized in Table~\ref{tbl:comparison-mate},  where ``+", ``-", and ``$\approx$" denote that a competitor performs better than, worse than, and similarly to ATM-R in terms of the IGD/HV value, respectively.

As shown in Table~\ref{tbl:comparison-mate}, ATM-R performed better than ATM-RMCDP, ATM-RMobj, and ATM-RMdiv in terms of both the IGD and the HV values. With regard to the IGD value, ATM-R was better than ATM-RMCDP, ATM-RMobj, and ATM-RMdiv on 19, 10, and 9 test functions, respectively. In contrast, ATM-RMCDP, ATM-RMobj, and ATM-RMdiv outperformed ATM-R on 5, 7, and 5 test functions, respectively. In terms of the HV value, ATM-R performed better than ATM-RMCDP, ATM-RMobj, and ATM-RMdiv on 22, 13, and 8 test functions, respectively. Inversely, ATM-RMCDP, ATM-RMobj, and ATM-RMdiv outperformed ATM-R on 3, 7, and 4 test functions, respectively. The experimental results show that the multiphase selection strategy is critical to ATM-R.

\section{Conclusions}
This paper has analyzed the key task of constrained multiobjective optimization in depth and decomposed it into three subtasks explicitly for the first time. To accomplish these three subtasks in different evolutionary phases, an adaptive tradeoff model with reference points (ATM-R) was designed. Specifically, ATM-R takes advantage of infeasible solutions to achieve different tradeoffs in these three subtasks. In the infeasible phase, ATM-R distinguishes and uses infeasible solutions with good diversity to enhance the diversity loss due to its pursuit of feasibility. Thus, the population can move toward feasible regions from diverse search directions. In the semi-feasible phase, ATM-R leverages infeasible solutions with good diversity/objective function values to promote the transition from ``the tradeoff between feasibility and diversity" to ``the tradeoff between convergence and diversity". Thus, the population can locate enough feasible solutions and approach the CPF quickly. In the feasible phase, ATM-R employs NSGAII to seek a set of well-converged and well-distributed solutions close to the constrained Pareto front. Moreover, a multiphase mating selection strategy is proposed to select appropriate mating parents adaptively. Experimental studies on a wide range of CMOPs demonstrate that:
\begin{itemize}
	\item 
	ATM-R achieves better or at least highly competitive performance against other representative CMOEAs.
	
	\item 
	ATM-R has a significant advantage over ATM for constrained multiobjective optimization.
	
	\item 
	The update mechanisms in the infeasible phase and the semi-feasible phase are both critical to the performance of ATM-R.
	
	\item The multiphase mating selection strategy is significant to the performance of ATM-R.
		
\end{itemize}
In the future, we will extend ATM-R to solve constrained expensive multiobjective optimization problems.

\begin{table}[t]
	\centering
	\caption{Results of ATM-RMCDP vs ATM-R, ATM-RMobj vs ATM-R, and ATM-RMdiv vs ATM-R on 36 test functions.}
	\begin{tabular}{ccc}
		\toprule
		Algorithms &\tabincell{c}{IGD\\ +/-/$\approx$}  & \tabincell{c}{HV\\ +/-/$\approx$} \\
		\midrule
		ATM-RMCDP vs ATM-R  & 5/19/12   & 3/22/11 \\
		ATM-RMobj vs ATM-R   & 7/10/19   & 7/13/16 \\
		ATM-RMdiv vs ATM-R   & 5/9/22    & 4/8/24  \\
		\bottomrule
	\end{tabular}%
	\label{tbl:comparison-mate}%
\end{table}%

\bibliographystyle{bib/IEEEtran}

\bibliography{bib/IEEEabrv,bib/mybib}

\end{document}